\def\paperID{36} % *** Enter the Paper ID here
\def\confName{3DV\xspace}
\def\confYear{2024\xspace}

\def\paperTitle{Generalizing Single-View 3D Shape Retrieval to Occlusions and Unseen Objects}

\def\authorBlock{
    Qirui Wu$^{1}$ \qquad 
    Daniel Ritchie$^{2}$ \qquad
    Manolis Savva$^{1}$ \qquad 
    Angel X. Chang$^{1,3}$ \\
    $^{1}$Simon Fraser University \qquad
    $^{2}$Brown University \qquad
    $^{3}$Alberta Machine Intelligence Institute (Amii)    
    \\
    {\tt\small \url{https://3dlg-hcvc.github.io/generalizing_shape_retrieval/}}
}

\newif\ifreview 
\newif\ifarxiv \newcommand{\arxiv}{\arxivtrue}
\newif\ifcamera 
\newif\ifrebuttal 

\arxiv
\pdfoutput=1
\documentclass[10pt,twocolumn,letterpaper]{article}
\ifreview \usepackage[review]{cvpr} \fi
\ifarxiv \usepackage[pagenumbers]{cvpr} \fi
\ifrebuttal \usepackage[rebuttal]{cvpr} \fi
\ifcamera \usepackage{cvpr} \fi

\usepackage{graphicx}
\usepackage{amsmath}
\usepackage{amssymb}
\usepackage{booktabs}

\usepackage{times}
\usepackage{microtype}
\usepackage{epsfig}
\usepackage[table,xcdraw]{xcolor}
\usepackage{caption}
\usepackage{float}
\usepackage{placeins}
\usepackage{color, colortbl}
\usepackage{stfloats}
\usepackage{enumitem}
\usepackage{tabularx}
\usepackage{xstring}
\usepackage{multirow}
\usepackage{xspace}
\usepackage{url}
\usepackage{subcaption}
\usepackage{xcolor}
\usepackage{comment}
\usepackage[hang,flushmargin]{footmisc}
\usepackage[numbers]{natbib}

\usepackage{custom_symbols}

\ifcamera \usepackage[accsupp]{axessibility} \fi

% ...

\newcommand{\mypara}[1]{\noindent\textbf{#1}}

\ifarxiv  \fi

\definecolor{tblblue}{RGB}{31, 119, 180}
\definecolor{tblorange}{RGB}{255, 127, 14}
\definecolor{tblgreen}{RGB}{44, 160, 44}
\definecolor{tblred}{RGB}{214, 39, 40}

\newcommand{\R}[1]{{%
    \textbf{%
        \ifstrequal{#1}{1}{\textcolor{tblred}{R#1}}{%
        \ifstrequal{#1}{2}{\textcolor{tblblue}{R#1}}{%
        \ifstrequal{#1}{3}{\textcolor{tblorange}{R#1}}{%
        \ifstrequal{#1}{4}{\textcolor{tblgreen}{R#1}}{%
                           \textcolor{cyan}{R#1}%
        }}}}%
    }%
}}

\expandafter\def\expandafter\normalsize\expandafter{%
    \normalsize%
    \setlength\abovedisplayskip{0pt}%
    \setlength\belowdisplayskip{0pt}%
    \setlength\abovedisplayshortskip{-8pt}%
    \setlength\belowdisplayshortskip{2pt}%
}  % Add packages to _macros.tex

\usepackage{xr-hyper}

\makeatletter
\newcommand*{\addFileDependency}[1]{
  \typeout{(#1)}
  \@addtofilelist{#1}
  \IfFileExists{#1}{}{\typeout{No file #1.}}
}

\makeatother

\usepackage[pagebackref,breaklinks,colorlinks]{hyperref}
\usepackage[capitalize]{cleveref}
\crefname{section}{Sec.}{Secs.}
\crefname{table}{Table}{Tables}
\crefname{figure}{Fig.}{Figs.}

\begin{document}

%% TITLE
\title{\paperTitle}
\author{\authorBlock}
\maketitle
\begin{abstract}
Single-view 3D shape retrieval is a challenging task that is increasingly important with the growth of available 3D data. Prior work that has studied this task has not focused on evaluating how realistic occlusions impact performance, and how shape retrieval methods generalize to scenarios where either the target 3D shape database contains unseen shapes, or the input image contains unseen objects. In this paper, we systematically evaluate single-view 3D shape retrieval along three different axes: the presence of object occlusions and truncations, generalization to unseen 3D shape data, and generalization to unseen objects in the input images. We standardize two existing datasets of real images and propose a dataset generation pipeline to produce a synthetic dataset of scenes with multiple objects exhibiting realistic occlusions. Our experiments show that training on occlusion-free data as was commonly done in prior work leads to significant performance degradation for inputs with occlusion. We find that that by first pretraining on our synthetic dataset with occlusions and then finetuning on real data, we can significantly outperform models from prior work and demonstrate robustness to both unseen 3D shapes and unseen objects. 
% \todo{Brainstorm the title.}
\end{abstract}
\section{Introduction}
\label{sec:intro}

\begin{figure}[t]
  \includegraphics[width=\linewidth]{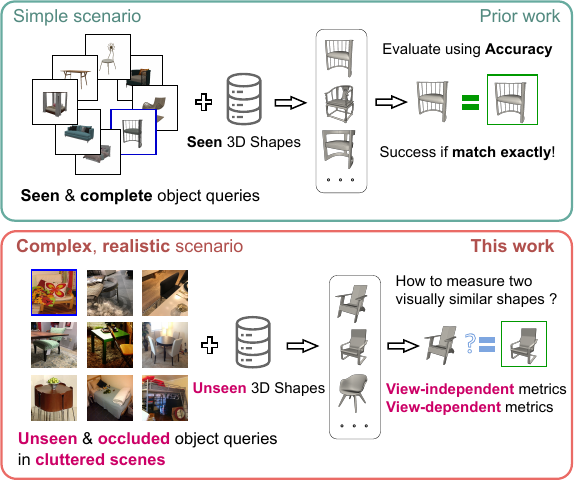}
  \vspace{-24pt}
  \caption{We focus on the single-view 3D shape retrieval task in complex but realistic scenarios where both the object and 3D shape candidates may be unseen during training, and the object may be observed in cluttered scenes under significant occlusions.}
  \label{fig:problem-state}
\end{figure}
% %------------------------------------------------------------------------

% Rapid advances in 2D perception have led to intelligent systems with accurate classification \todo{cite}, object detection \todo{cite} or segmentation \todo{cite} of cluttered, real-world images. Although they are well formulated in 2D domain, 
3D shape retrieval given a single-view image is increasingly useful for 3D content creation, robotic perception and other applications.
The increasing volume of 3D data for domains such as e-commerce, interior design, and augmented reality makes single-view shape retrieval an important task.
However, this is a challenging task.
Paired real image--3D shape data require significant annotation so there are no large-scale, diverse datasets.
Synthetic data can be helpful in bridging this gap but prior work has deployed it in simplified, unrealistic settings where a single synthetic 3D object is overlaid on a blank or random image background (see \Cref{fig:problem-state}).
% The difficulty in single-view 3D shape retrieval is underestimated in a way that prior works \todo{cite} mostly evaluate their approaches under a simpler setting where a single object is rendered from synthetic dataset \todo{cite} with blank background.
Some work leverages existing, relatively small image-shape datasets such as Pix3D~\cite{sun2018pix3d} to demonstrate retrieval performance.
However, results are not comparable due to the lack of a standardized, shared evaluation protocol, leaving work to create its own training and evaluation splits, and pick its own metrics for evaluation.
% \angel{emphasize that the main issue is the lack of a proper evaluation protocol}
Lastly, occlusion or truncation of objects in the image are prevalent scenarios that impact task difficulty in real data but are usually ignored~\cite{lin2021single}.

In this paper, we systematically study the generalization of single-view 3D shape retrieval methods along three axes, making associated contributions in each axis.

\mypara{Generalization to unseen objects.}
%Though some prior work does retrieval given novel 3D shapes (i.e. unavailable at training time)~\todo{cite}, 
Prior work on shape retrieval has not been evaluated in settings involving unseen objects (i.e. no overlap in 3D shapes between train and val/test splits).
Thus, it is unclear how existing retrieval methods generalize to images containing unseen objects.
We present experiments that evaluate on this setting, and demonstrate that this form of generalization to unseen objects remains challenging (see \Cref{tab:pix3d_easy_hard} for a summary of performance degradation in the unseen object setting).

\mypara{Generalization to occluded objects.}
% Given an real photo of cluttered complex scene, it is common that objects are partially observed because of occlusion or truncation. Prior works \todo{cite} agree that occlusion leads to worse retrieval results, but there is no thorough analysis of to what extent occlusion degrades shape retrieval.
Occlusion and truncation occur in many real images but prior work has not studied their impact during training and in evaluation.
We construct a dataset with varying occlusion rates and systematically analyze the impact of occlusions on retrieval performance.
% And there is no answer based on experiments to the question whether occlusion really matters when directly training with occluded objects. Some works \todo{cite} first predict target object masks using pretrained 2D segmentation methods~\cite{he2017mask, cheng2022masked}. However, how the quality of masks influence the image feature extraction is not validated. Those masks are either too coarse to capture geometry details of objects in images or incomplete since objects can be occluded or truncated by other stuff in the scene.

%\mypara{Generalization to novel 3D shape database.}
\mypara{Generalization to similar shapes.}
Most work on image-to-shape retrieval has been on evaluated on small shape databases.  There has been no systematic study of performance of models when tested on large databases where there may be \emph{many} shapes that are similar to the one annotated GT shape.  In such cases, it is possible that the common accuracy metric fails to capture whether more similar shapes are ranked higher than dissimilar shapes.
While reconstruction metrics such as Chamfer distance and volumetric intersection-over-union (IoU) have been used for evaluation, these metrics are limited in evaluating fine-grained distinctions between object instances in a robust way.
We propose a suite of view-dependent and view-independent metrics that are more appropriate for this practical setting.
% However, since there is no ground truth corresponding 3D shape, metrics like chamfer distance and IoU are used to show performance. However, these reconstruction-driven metrics are limited in revealing how good retrieved shape match local structure details of object in image as objects in the same category could share the same or similar overall geometry.

In summary, we make the following contributions: 1) we propose a standardized evaluation protocol that better characterizes 3D shape retrieval performance and we recommend a suite of view-dependent and view-independent metrics; 2) we develop a multi-object occlusion dataset to measure the impact of occlusions on retrieval; 3) we experimentally analyze the generalization of 3D shape retrieval methods on unseen objects and novel 3D shape databases.

\begin{table}
\centering
% \resizebox{\linewidth}{!}
{
\begin{tabular}{@{}ll rrrr@{}}
\toprule
Dataset & objects & $Acc_1\uparrow$ & $Acc_5\uparrow$ & $\text{CD}\downarrow$ & $\text{CD}_5\downarrow$  \\
\midrule
\multirow{2}{*}{Easy set} & seen & 74.9 & 89.2 & 0.35 & 1.66 \\
% & unseen & 0.28 & 7.58 & 1.5222 & 1.7735  \\
& unseen & 0.3 & 7.6 & 1.52 & 1.77  \\
% \multirow{2}{*}{Hard set} & seen & 47.2 & 73.53 & 0.9941 & 2.0423 \\
\multirow{2}{*}{Hard set} & seen & 47.2 & 73.5 & 0.99 & 2.04 \\
% & unseen & 0.00 & 17.4 & 2.3984 & 2.6206 \\
& unseen & 0.0 & 17.4 & 2.40 & 2.62 \\
\bottomrule
\end{tabular}
}
\vspace{-8pt}
\caption{
Performance of our CMIC~\cite{lin2021single} implementation trained on the Pix3D ``Easy set'' as reported by \citet{lin2021single}, and tested on val sets of Pix3D that contain unobserved objects (``unseen'') or contain occlusions (``hard set'').
Note the significant performance degradation across metrics.
}
\label{tab:pix3d_easy_hard}
\end{table}

% \begin{table}
% \centering
% \resizebox{\linewidth}{!}
% {
% \begin{tabular}{@{}l rrrr rrrr@{}}
% \toprule
% & \multicolumn{4}{c}{Easy Set} & \multicolumn{4}{c}{Hard Set} \\
% \cmidrule(lr){2-5}\cmidrule(lr){6-9}
% & $Acc_1$ & $Acc_5$ & $\text{CD}$ & $\text{CD}_5$ & $Acc_1$ & $Acc_5$ & $\text{CD}$ & $\text{CD}_5$  \\
% \midrule
% Easy Set & - & - & - & - & - & - & - & -  \\
% Hard Set & - & - & - & - & - & - & - & -  \\
% All Set & - & - & - & - & - & - & - & -  \\
% \bottomrule
% \end{tabular}
% }
% \caption{\todo{add caption. For metrics, use what Pix3D use for comparison.}}
% \label{tab:pix3d_easy_hard}
% \end{table}

\section{Related Work}
\label{sec:related}

3D shape retrieval is well studied~\cite{tangelder2008survey,li2015comparison}.
Task variants involve querying of shapes from single-view images~\cite{li2015joint,lin2021single}, sketches~\cite{he2018triplet}, text, as well as partial scans~\cite{hua2017shrec,pham2018shrec} and other 3D objects~\cite{bai2016gift,he2018triplet}.
Here, we describe relevant work focusing on single, photo-realistic image to 3D shape retrieval.

% shape to shape retrieval, or sketch to shape retrieval \cite{he2018triplet} proposes enhanced triplet-center loss but only demonstrates on synthetic renderings instead of real images. 

\mypara{Image to shape retrieval.}
% There have been growing interest in developing algorithms of 3D shape retrieval from a single image. 
% A prescribed approach~\cite{chen2003visual} represent 3D shapes as LFD descriptors and retrieve target shape by comparing the similarity of obtained features. Compared to LFD, Lee et al.~\cite{lee2018cross} develop a learning-based method to handle the image-based shape retrieval task. 
3D shapes are commonly represented as multiview image renderings~\cite{li2015joint,massa2016deep, grabner2019location, kuo2020mask2cad, lin2021single, fu2020hard} to shrink the gap between 2D and 3D.  
%By creating joint-embeddings of shapes as multi-view images, such methods can be used for image-to-shape retrieval, as well as shape-to-shape~\cite{li2015joint,he2018triplet} or sketch-to-shape retrieval~\cite{he2018triplet}.
% As all approaches tackle with synthetic 3D data, there exist domain shift from query images and shape renderings. 
To align the different modalities of the same physical entity, real RGB images and shape multiview images are encoded into a joint-embedding space using triplet or contrastive losses.
There is also recent interest in studying generalization to novel objects and occlusions.
\citet{nguyen2022templates} use templates for object pose estimation, and study the robustness of 3D shape retrieval to occlusions in the object pose estimation task.

Early work by \citet{li2015joint} used a joint embedding space of 3D shapes and 2D images to retrieve shapes from images.
\citet{lee2018cross} proposed to learn a joint embedding space for natural images and 3D shapes in an end-to-end manner. %The triplet loss is used to joint optimize image and shape embedding, encouraging embeddings of data with the same entity to get closer than those with different entities.
% \citet{grabner2019location} leveraged of location fields in this task. %that possess rich information but are position sensitive.
To handle different texture and lighting conditions in images, \citet{fu2020hard} proposed to synthesize textures on 3D shapes to generate hard negatives for additional training.
% Some work does not rely on prerendered multiview images but instead first predicts the 3D object pose and then renders depth images with the estimated pose~\cite{grabner20183d}.
%Instead of digesting multiviews, \cite{grabner20183d} first predicts 3D object pose with a PnP algorithm and then renders depth images from 3D shapes under the estimated pose. 
\citet{uy2021joint} used a deformation-aware embedding space so that retrieved models better match the target after an appropriate deformation.
A recent model, CMIC~\cite{lin2021single} achieved state-of-the-art performance on retrieval accuracy by using instance-level and category-level contrastive losses.
We build on CMIC for our analysis of image to 3D shape retrieval generalization.
% Most of previous works discussed how to improve the accuracy of shape retrieval by learning some critical image/shape features or reasonable embedding space with metric learning.
% CMIC~\cite{lin2021single} achieves the-state-of-art performance on several image-based shape retrieval benchmarks by designing instance-level and category-level contrastive losses.

\mypara{Joint retrieval and alignment.}
Another line of work uses image-to-shape retrieval as a component in an image-to-scene pipeline~\cite{gupta2015aligning,izadinia2017im2cad, kuo2020mask2cad, gumeli2022roca}, where objects are detected, segmented, aligned, and CAD models are retrieved and posed to form a 3D scene.
In these works, the layout estimation and CAD model alignment tasks are often solved jointly.
IM2CAD~\cite{izadinia2017im2cad} proposed a pipeline that produces a full indoor scene (room and furniture) from a single image by leveraging 2D object recognition methods to detect multiple object candidates. %At the CAD model selection stage, they leverage the fine-tuned Faster-RCNN ~\cite{ren2015faster} network to compute features of detected image bounding box and each of the shape renderings. Different from the prior methods, this pipeline is in the context of multi-objects in the input RGB image, so it integrates 2D instance segmentation. And it additionally predicts the 3D pose. 
Mask2CAD~\cite{kuo2020mask2cad} proposed a more lightweight approach that jointly retrieves and aligns 3D CAD models to detected objects in an image. Patch2CAD~\cite{kuo2021patch2cad} used a patch-based approach to improve shape retrieval.
ROCA~\cite{gumeli2022roca} and its follow-up~\cite{langer2022sparc} relied on dense 2D-3D correspondences to retrieve geometrically similar CAD models but mainly focused on differentiable pose estimation.
Point2Objects~\cite{engelmann2021points} addressed the same task as Mask2CAD but directly regresses 9DoF alignments and treats object retrieval as a classification problem, and mainly showed results on synthetic renderings.

\mypara{Evaluating image to shape retrieval.}
To study pose estimation, researchers have created datasets with 3D shapes aligned to real-world images~\cite{xiang2014beyond,xiang2016objectnet3d}. However, these datasets are not suitable for shape retrieval as the shape may not be an accurate match.
\citet{li2019shrec} provided a benchmark on monocular image to shape retrieval with 21 categories. Their images mostly offered unoccluded views of the object and the shape match was inexact (category match, shape geometry not guaranteed to match).
There are few datasets with accurate image-shape match, leading to relatively small-scale and ad-hoc evaluation schemes in prior work.
% Challenging to create some datasets as it is difficult to find exact shapes that match object in real-world images.  The annotation process is time-consuming and typically consists of restricting to well-known products with both image and shape, starting with the shape and finding real-world image matches, or building custom dataset of image and scanned shape.
\citet{li2015joint} evaluated on a small dataset of 105 shapes paired with 315 google search images (each shape had 3 images), with the 105 shapes excluded from training.
The authors noted that this benchmark was expensive to create.
\citet{sun2018pix3d} contributed Pix3D, a dataset of images aligned to posed 3D shapes and demonstrated its usefulness for benchmarking methods for single-view 3D reconstruction, image-to-shape retrieval, and pose estimation.
The Pix3D dataset provided annotations of whether the shape in the image had truncation or occlusion.
However, experiments in this work and followups~\cite{sun2018pix3d,fu2020hard,grabner2019location,lin2021single} are limited to untruncated and unoccluded objects, and few work study generalization to unseen shapes~\cite{grabner2019location,lin2021single} in a rigorous manner.
MeshRCNN~\cite{gkioxari2019mesh} proposed standard splits for Pix3D, including a more challenging data split S2 for testing on images of unobserved objects, but this split was rarely used in followup work.
Mask2CAD~\cite{kuo2020mask2cad} was one of the few works that used the S2 data split.
%and \citet{grabner2019location} evaluated retrieval performance with unseen ShapeNet models. 
We standardize two real image datasets, and develop a new synthetic dataset and associated metrics to evaluate 3D shape retrieval in a more %standardized and 
comprehensive manner than prior work.

\section{Analyzing image-to-shape retrieval}
\label{sec:overview}

\begin{figure}[t]
\includegraphics[width=\linewidth]{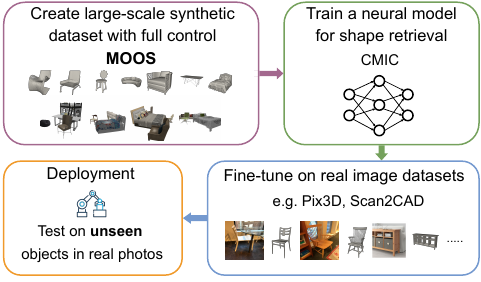}
\vspace{-20pt}
\caption{
Overview of our single-view 3D shape retrieval training pipeline.
We pretrain with a large synthetic dataset exhibiting object occlusions.
After fine-tuning on real image datasets we significantly improve generalization to occlusions and previously unseen objects.
}
\label{fig:high_level_diagram}
\end{figure}

%\angel{We should describe the overall evaluation setup with the components and what we aim to achieve before jumping into the model}

% \angel{clarify the role of the mask in our problem statement, do we assume it is given?}
To study the performance of image-to-shape retrieval, we argue that it is important to: 1) evaluate on unseen shapes by establishing a train/val/test split that ensures separation of val/test shapes not seen during training; 2) evaluate on occluded and truncated objects; and 3) evaluate the geometric similarity of the ranked results to the ground truth shape.
%In a realistic deployment of a image to 3D shape system, we would likely have a fixed set of aligned image-shape data for training a retrieval model, which is then deployed on novel images, as well as a large, potentially growing dataset of 3D shapes. 
In addition, we believe it is important to evaluate on real-world images while still
leveraging synthetically constructed scenes and rendered images for controlled experiments and pre-training (see \Cref{fig:high_level_diagram}).
To this end, we propose an evaluation protocol with standardized train/val/test splits for three datasets, two of which have 3D shapes aligned to real-world images, and a third that is synthetically constructed.
For each dataset, we ensure that the val/test splits contain shapes not seen during training.
As part of our evaluation, we also analyze existing 3D similarity evaluation metrics and propose a set of recommended metrics for image-to-shape retrieval.

We select Pix3D~\cite{sun2018pix3d} as our first evaluation dataset as it provides accurate 3D shape-to-image matches and annotations indicating occlusion and truncation.
Our second dataset is ScanIm2CAD: a new dataset using images from ScanNet~\cite{dai2017scannet} with projected masks and ShapeNet~\cite{chang2015shapenet} models from Scan2CAD~\cite{avetisyan2019scan2cad} annotations.
We create a standardized image-to-shape retrieval benchmark on this dataset as it is a popular choice for recent work on joint shape-retrieval and alignment~\cite{kuo2020mask2cad,kuo2021patch2cad,gumeli2022roca,langer2022sparc}.
The 3D shape to image match is not as high as Pix3D but this dataset offers real-world images of objects in context (in real-world scenes), with realistic occlusions and truncations.
Lastly, we create \moos, a dataset of synthetic scenes where we programmatically control the selection of objects and layouts.
\moos allows us to easily render images with ground-truth masks that are accurately aligned with ground-truth CAD models, and control the amount of occlusion and truncation.

Using these datasets, we investigate how well a state-of-the-art image-to-shape retrieval model, CMIC~\cite{lin2021single}, can generalize to unseen shapes, and handle images with occluded and truncated objects.
We find that performance drops drastically under these challenging settings (see \Cref{tab:pix3d_easy_hard}).
We use our synthetically generated \moos dataset to systematically investigate the impact of occlusion and unseen shapes (see \Cref{sec:results}) and show that pretraining with \moos improves performance on real-world images (see \Cref{tab:transfer_from_syn_to_real}).
%For our experiments, we select a state-of-the-art method, CMIC~\cite{lin2021single} as the retrieval model of choice.

% In the rest of this paper, we describe in detail the image-to-shape model (\Cref{sec:model}), datasets we use (\Cref{sec:dataset}), metrics (\Cref{sec:metrics}), and our experimental results and findings (\Cref{sec:results}).

\section{Model}
\label{sec:model}

% \angel{clarify whether we are working with the entire image or there is a object mask that indicates the object of interest.  Clarify whether a image can contain multiple objects.}
% \qirui{The original image can contain multiple objects and we use object mask to indicate the object of interest. We crop the image and mask using 2D bbox computed from object mask. Thus the resized query image and mask passed into the query encoder should mainly contain one object with potential occlusions.}

We use CMIC~\cite{lin2021single} as the basis of our benchmarking.
CMIC learns a joint embedding space between images and 3D shapes using contrastive losses at the instance and category levels.
Given an image containing a target object, the joint embedding space is used to retrieve a ranked list of 3D shapes based on similarity to the visual features of the object in the image.
Since the input images may contain multiple objects we use an object mask to indicate the object of interest.
The image and mask are cropped from the input image using the 2D bounding box computed from the mask. As input to our models, we use the resized cropped image and mask and assume that the crop mainly contain only the object of interest, potentially with occlusions.
%that best matches the geometry of an object in the image. %with the maximum cosine similarity between them in the embedding space. 
% Figure~\ref{fig:image-encoder-arch} illustrates the architecture of our method.

\mypara{Encoders.} 
CMIC consists of two separate encoders to obtain representations $f^I$ and $f^S$ for query image $I$ and 3D shape $S$.
% As query images depict real-world cluttered indoor environments, the image encoder could be misled by information unrelated to the target object.
The object segmentation mask $M$ is fed to the query encoder to mask out background information.
% \qirui{Here I try to state that we use object mask to indicate the object of interest}
Each 3D shape $S$ is represented as a set of multiview images $\{S_k\}_{k=1}^m$ rendered from predefined camera viewpoints.
This approach aims to bridge the gap between the 2D and 3D modalities by converting the image-shape retrieval problem to image-image retrieval.
By passing $m$ images of one shape into the shape encoder, we obtain a set of image features $\{f^{S_k}\}_{k=1}^m$.
For each shape in a batch, we compute multiple query-conditioned features $\{f^{S_k}_i\}_{i=1}^B$ for all queries $\{I_i\}_{i=1}^B$ using dot product attention.

% \subsection{Image Encoder.} Shown by \Cref{fig:image-encoder-arch}, we use ResNet-50 feature pyramid network ~\cite{lin2017feature} to extract feature map from the given RGB image crop. To be specific, the input image crop is $224 \times 224$ and we use the $C_{4}$ layer output ($256\times14\times14$) as the encoding result (ROI feature). The level of the layer is selected by:
% \begin{align}
% k=\left\lfloor k_{0}+\log _{2}(\sqrt{w h} / 224)\right\rfloor
% \end{align}
% where $k_{0}$ is set to be 4. 

% \subsection{Image Decoder.} We take the ROI feature as the input and use the same decoder architecture used by ~\cite{kuo2020mask2cad}, which is a stack of $3\times3$ convolution layers. See \Cref{tab:network_arch_of_decoder} for details.

% \subsection{Shape Encoder.} Shown by \Cref{fig:shape-encoder-arch}, we use PointNet++~\cite{qi2017pointnet++} as the shape encoder backbone. To be specific, we only leverage the classification branch as our task is closer to 1-of-\(K\) classification task and we use the output vector of PointNet on that branch to be the encoder output, where $C_{4}$ is 1024. Pix3D dataset only provides object mesh, so the input point cloud is obtained in a pre-process in which we uniformly samples 2048 points from the 3D surface based on the triangle area.

\mypara{Image-Shape Joint Embedding.}
Given query features $\{f^i\}_{i=1}^B$ and query-attended shape features $\{f^{S_k}_i\}_{i=1}^B$, we learn an image-shape joint embedding using contrastive losses. 
% The representations from different modalities pull each other when 2D object and 3D shape matches and push away if they don't match.
% \subsection{Image-shape joint embedding space.} 
% Mathematically, we denote the input RGB crop as \(I\) and binary mask crop as \(M\). We define the image encoder as a function: \(I \circ M\rightarrow f^{im}\), where \(f^{im} \in \mathbb{R}^{d}\). We compare image embedding $f^{im}$ with shape embeddings $f^{obj}$ in the joint embedding space, which we will discuss later. We denote the underlying shape id of \(i\)-th shape embedding as \(s_{i}\). 
We denote $D(f^{i}, f^{S_k}_j)$ as the similarity function between the embeddings of the $i$-th image and the $k$-th query-specific shape rendering of the $j$-th image.
% , where \(D\) is the similarity function with a temperature hyperparameter $\tau$: 
% \begin{align}
%     D(x, y):=e^{\frac{1}{\tau}\left(\frac{x}{\|x\|}\right)^{T}\left(\frac{y}{\|y\|}\right)}
% \end{align}
% The core of our approach is learning to map between 2D image observation of an object and 3D CAD models. We formulate this problem into an image-shape embedding space learning problem. To do this, we use a \pccrop architecture that is adapted from Mask2CAD~\cite{kuo2020mask2cad} network to construct image-shape embedding space (see \ref{fig:network_overview}). This architecture uses a ResNet~\cite{he2016deep} feature pyramid network~\cite{lin2017feature} as the image encoder backbone, and a PointNet++~\cite{qi2017pointnet++} network as the shape encoder backbone. The decoder is a stack of \(3\times3\) convolution layers on the masked ROI features; Table~\ref{tab:network_arch_of_decoder} shows the detailed architecture of it.  We refer to the resulting extracted features descriptors as $f^{im}$ and $f^{obj}$ for the image and shape, respectively. 
The contrastive losses are designed at two levels: instance and category.
The instance-level contrastive loss treats matching image-shape pairs as positive examples, and all other cases as negative.
Thus, each image $I_i$ is paired with only one positive shape $S_i$ and $B-1$ negative shapes $S_j$ where $j \in B \setminus \{B_i\}$: %The instance contrastive loss is expressed in the following form: 
% \begin{align}
\[
    L_\text{inst}=-\sum_{i \in B} \log \frac{D\left(f^{i}, f_i^{S_i}\right)}{\sum_{j \in B}D\left(f^{i}, f_j^{S_k}\right)}
\]
% \end{align}
% where \(f_{p}^{o b j}\) represents the feature descriptor of a corresponding 3D object to the image region, \(f_{n}^{o b j}\) the feature descriptor of a non-corresponding object, \(C\) a weighting parameter. For each positively corresponding object (determined by the CAD annotation to the image), \(N\) represents the set of negatively corresponding objects, which is composed of other CAD models in the training batch. Since our problem statement assumes the category of the object is known; thus, our embedding spaces are constructed for each object category and the weights are shared among them.

The category-level contrastive loss is used to cluster and disperse image-shape pairs belonging to the same and different categories.
This loss leverages shape category labels $y_i$ to maximize category agreement in a batch: % which can be formulated as follows:
% \begin{align}
\[
    L_\text{cat}=-\sum_{i \in B} \frac{1}{|C(i)|} \sum_{c \in C(i)} \log \frac{D\left(f^{i}, f_i^{S_c}\right)}{\sum_{j \in B}D\left(f^{i}, f_j^{S_k}\right)}
\]
% \end{align}
where $C(i)$ refers to all instance with the same category label (e.g. $\{j|j\in B \setminus \{B_i\} \text{ and } y_j=y_i\}$). There could be multiple positive and negative samples for an image in the same mini-batch. The total loss is a weighted sum of the instance and category contrastive losses: $L=L_\text{inst} + \beta_1 \cdot L_\text{cat}$ where $\beta_1$ is the weight on the category loss.

% Note that in Mask2CAD~\cite{kuo2020mask2cad} paper, it is called distance function but we correct it by using similarity here because the closer vectors get higher output value from that function. We can also confirm that by the loss function where it optimizes towards the higher \(D\) function output for positive (image, shape) embedding pair and lower \(D\) function output for the negative pairs.

\mypara{3D Shape Retrieval.}
The multiview image features of all 3D shape candidates are computed offline.
At inference time, an RGB query is embedded using the query encoder.
A list of 3D shapes is retrieved by ranking based on the cosine similarity between the query-conditioned shape feature and the query feature in descending order.
We assume the category label is known to retrieve appropriate 3D shapes.
% At test time, the input is only a single-view RGB image cropped by the bounding box, a segmentation mask and the object category. We use the image-shape embedding space to compute distances between the computed image embedding from input image and all other shape embeddings in this space. We pick \(N_{k}\)-nearest neighbor shapes as our retrieval result and sort this list by the value of the distance function. It might be a challenge when the number of candidate shapes grows large since no matter which \(N_{k}\) we choose, we have to sort the whole list of candidate shapes first and then pick the top-\(N_{k}\).
% We then sort shapes by the that distance in ascending order and the output is \(\{s_{i_{1}}, s_{i_{2}}, \ldots, s_{i_{N_{k}}}\}\) where \(D\left(f^{i m}, f_{i}^{s_{i_{1}}}\right) \leq D\left(f^{i m}, f_{i}^{s_{i_{2}}}\right) , \ldots, \leq D\left(f^{i m}, f_{i}^{s_{i_{N_{k}}}}\right)\), assuming we return top-\(N_{k}\) shapes.

%------

\mypara{Implementation.}
We use a pretrained ResNet50~\cite{he2016deep} to encode RGB images with object masks as an extra channel and a pretrained ResNet18~\cite{he2016deep} to encode grayscale multiview renderings of 3D shapes.
The extracted image/rendering embeddings are projected to features of dimension 128 with additional MLPs.
% To accommodate the inputs including image and object mask, the first convolution layer of ResNet50 is adjusted by adding one extra channel for transforming binary mask information. We extract the output from the last fully-connected layer as the image feature of dimension 128. For the shape encoder, we instead choose ResNet18 pretrained on ImageNet to embed grayscale multiview renderings of dimension 128, which means the number of channels of the first convolution layer is modified from 3 to 1. 
We render 12 images for each 3D shape using predefined camera poses placed on the same horizontal plane and 30 degrees apart from each other.
We apply several data augmentations to image queries including affine transformation, crop, flip and color jitter.
We use temperature $\tau=0.1$ and $\beta_1=0.2$.
We train the model on a single Nvidia A40 GPU with a batch size of 64 using an Adam~\cite{kingma2014adam} optimizer with initial learning rate $5e^{-5}$ and $\text{betas}=(0.5, 0.999)$.

\section{Datasets}
\label{sec:dataset}

Most work on image-to-shape retrieval rely on datasets of image-shape pairs \cite{xiang2014beyond, xiang2016objectnet3d, wang20183d, sun2018pix3d, avetisyan2019scan2cad} are constructed from realistic photos and synthetic 3D shapes.
Due to the difficulty of manually aligning 3D shape to 2D objects, existing datasets are limited in data volume, both in terms of 2D images and 3D shapes for various categories.
Unlike Pix3D whose accurate image-shape matches are obtained by using the IKEA object name or 3D scans of objects by the authors, other datasets~\cite{avetisyan2019scan2cad} provide 3D shapes that do not fully match the geometry of the observed objects.
Moreover, the annotated 3D object poses are subject to errors, which makes robust evaluation of single-view 3D shape retrieval methods harder.
We propose a scalable and simple synthetic data generation pipeline for constructing 3D scenes from sampled 3D shapes that mimic realistic object occlusions.

% 3DFuture steps forward by leveraging semi-realistic renderings of artist-designed 3D houses as image queries and CAD annotations are naturally obtained from scene construction process. 

\begin{table}
\resizebox{\linewidth}{!}
{
\begin{tabular}{@{} l rrrrrr@{}}
\toprule
Dataset & \#images & \#shape & \#cat & alignment & scalable & occ info \\
\midrule
% ObjectNet3D~\todo{cite} & - & - & - & - & - \\
Pix3D~\cite{sun2018pix3d} & 10K & 395 & 9 & accurate & no & partial \\
Scan2CAD~\cite{avetisyan2019scan2cad} & 25K & 3,049 & 35 & inaccurate & no & no \\
% 3DFuture~\todo{cite} & - & - & - & - & - \\
\midrule
MOOS & 120K & 6,209 & 4 & accurate & yes & complete \\
\bottomrule
\end{tabular}
}
\vspace{-8pt}
\caption{Comparison of datasets with image and 3D shape pairs. Note that Pix3D only categorizes objects into three groups based on the degree of object occlusions.
% \qirui{For \# of unique shapes, should I post numbers for all categories or just 4 categories we are interested?}
}
\label{tab:moos_stats}
\end{table}

\subsection{Real Image Datasets}

We evaluate on two datasets with 3D shapes aligned to real images, Pix3D~\cite{sun2018pix3d} and Scan2CAD~\cite{avetisyan2019scan2cad}.
Following prior work~\cite{lin2021single, fu2020hard}, we conduct experiments on 4 categories, chair, bed, sofa and table. There are $8,650$ images and $324$ unique 3D shapes in Pix3D over the $4$ categories. We create two sets using Pix3D annotations: \easy and \hard, where the latter contains occluded or truncated objects. We intentionally exclude 1,233 images of 181 objects from the training set to test generalization to unseen objects. For the remaining 7,417 images of 143 seen objects, we split images for each object into train and val sets in a 75:25 ratio. The Scan2CAD dataset collects scan-to-shape annotations by aligning 3D shapes from ShapeNet to 3D scans from ScanNet~\cite{dai2017scannet}. We use the ScanNet25K frames~\cite{dai2017scannet} as images and project 3D shapes onto each image using the camera poses. We use the ScanNetv2 splits resulting in 40K/11K image queries in train/val with 1,779 unique shapes over the
four categories. %chair, bed, table and sofa categories.

\begin{figure}
\includegraphics[width=\linewidth]{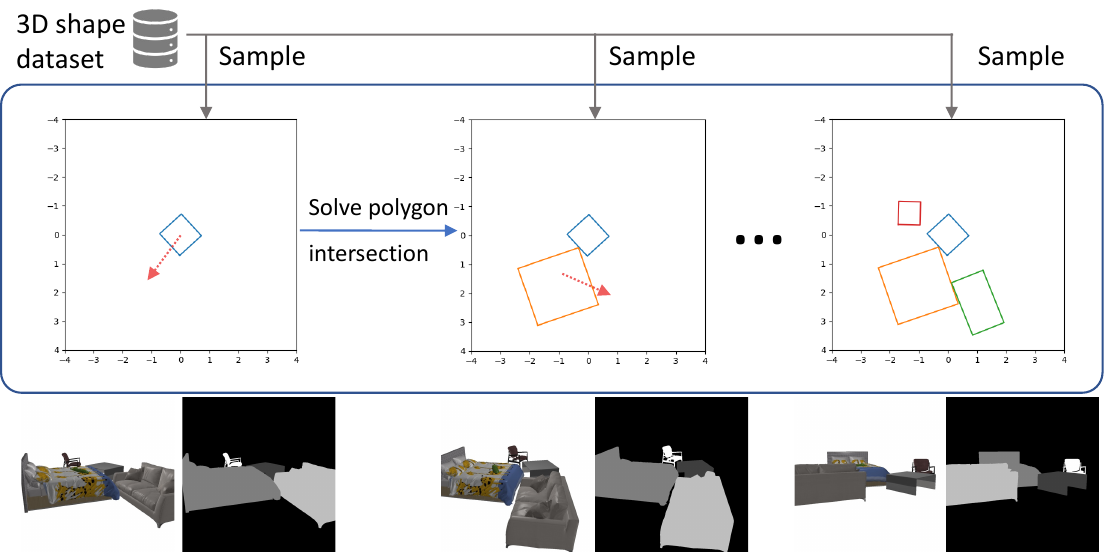}
\vspace{-20pt}
\caption{The pipeline of our Multi-Object Occlusion Scenes (MOOS) dataset generation procedure.
We sample and iteratively place objects while avoiding collisions.
The resulting scenes contain a variety of object categories and exhibit realistic occlusions.
}
\label{fig:moos_pipeline}
% \caption{We only use the $C_{4}$ layer output as encoder output in the ResNet-50-FPN architecture. The PointNet++ has both segmentation and classification branch and we only use the latter part of network. Output vector dimension $C_{4}$ is 1024 in our experiment.}
\end{figure}

\subsection{\moos: Multi-Object Occlusion Scenes}

To facilitate studying generalization in single-view shape retrieval, we propose a scalable synthetic dataset generation pipeline that we call Multi-Object Occlusion Scenes (\moos).
% ~\footnote{We will open-source both the generation pipeline and the dataset.}.
Our generation pipeline allows for control over the key variables we study: 1) amount of occlusion and truncation; and 2) novel shapes that are set aside for evaluation and not seen in training.  Compared to prior datasets, \moos provides a scalable dataset with accurate image-to-shape alignment and occlusion statistics (see \Cref{tab:moos_stats}).

Scenes in MOOS are comprised of randomly selected 3D shapes from 3D-FUTURE~\cite{fu20213d} that are randomly arranged to form a scene.
Compared to ShapeNet, furniture objects in 3D-FUTURE have more consistent geometric quality, higher-resolution textures, and known physical dimensions.
% ~\footnote{We find a small number of objects to be inaccurately sized and discard those.}.
% With the original object sizes, we can more accurately capture occlusion relationship among different objects.
Each generated scene consists of 4 objects from 4 categories (chair, bed, table and sofa), with the same shape potentially occuring in multiple scenes. To generate a scene layout with natural occlusions in real environment, we use a heuristic algorithm that iteratively places newly sampled 3D shapes into the existing layout, ensuring the 2D bounding box from a top-down view does not intersect with previously inserted 3D shapes. See \Cref{fig:moos_pipeline} for an overview. In this way, we compose a scene with closely placed objects exhibiting natural occlusion patterns, but no inter-penetrations.
See supplement for more details on the generation procedure.

Using our pipeline we generate 10,000 unique scenes to construct the \moos dataset.  With our generated scenes, we construct a dataset of rendered images with occluded shapes for retrieval (see \Cref{tab:moos_stats} and supplement for detailed statistics).
%Each shape may occur multiple times in different scenes.
%See the supplement for more detailed statistics.
We render 12 viewpoints per scene by evenly dividing along the azimuth every 30 degrees and sampling elevation uniformly between 5 and 25 degrees. We render an RGB image, instance segmentation, depth map, normal image and object-level RGB and mask images.
Thus, each scene in MOOS has 156 renderings, all at 1K$^2$ resolution.
The entire dataset consists of 1,560,000 images for the 10K scenes and is generated using PyTorch3D~\cite{ravi2020accelerating} in 35 hours.
As truncation is a special case of occlusion, we compute the occlusion rate for each object instance by comparing its intact object mask and instance mask.
Based on occlusion rate, we separate all object instances in MOOS into two subsets, the \emph{Occ} set and the \emph{NoOcc} set with 351K and 118K queries respectively, depending on whether occlusion exists.
Both subsets are split into train, val and test splits in an 8:1:1 ratio.
To investigate how shape retrieval methods generalize to unseen objects, we set aside $10\%$ of the objects (randomly selected) and their corresponding images from the train split.  
%\Cref{tab:moos_stats} contrasts \moos with datasets from prior work.

\section{Metrics}
\label{sec:metrics}

% \subsection{Metrics}

Typical shape retrieval metrics require both the ground-truth (GT) shape and suggested 3D shapes to compute either accuracy or point cloud based reconstruction scores~\cite{lin2021single, kuo2020mask2cad}, such as Chamfer Distance (CD), Normal Consistency (NC) and $\mathrm{F}1^{t}$.
% conditioned on distance $t$ that indicates the strictness of points being accurately reconstructed. 
\cite{kuo2020mask2cad, kuo2021patch2cad} also adopt $\text{AP}^{\text{mesh}}$ from~\citet{gkioxari2019mesh} to compute average precision weighted by recall at different IoU thresholds based on $\mathrm{F} 1^{0.3}$ scores.
These metrics are view-independent in a way that ignores potential occlusions in real images. 
Moreover, reconstruction metrics are sensitive to the point sampling method and the number of sampled points.
It is also unclear whether they robustly reflect how close retrieved shapes are to the ground truth.
We perform quantitative and qualitative analysis (see supplement for details) to select appropriate metrics for shape retrieval. 
We summarize the details of the metrics we chose here. 
% and divide them them into three groups: 1) accuracy metrics, 2) shape-wise metrics, and 3) view-dependent metrics. 

% \subsection{Retrieval and Shape-wise Metrics}
\subsection{View-independent Metrics}

% \textbf{Accuracy.}
% For query images with collected 3D shape annotations, retrieval category accuracy (\textit{\catacc}) and 3D shape accuracy (\textit{\acc}) are two direct metrics to measure the retrieval performance. 
We use Accuracy (\textit{\acc}) and Category Accuracy (\textit{\catacc}) to measure the average number of retrieved objects with correct shapes and categories, respectively. 
$Acc_k$ indicates whether the top-K retrieved objects contain the GT 3D shape.
However, accuracy metrics do not measure whether retrieved shapes are structurally similar to the GT, but not exact matches.
% However, accuracy metrics are limited in only considering the matchness of the target shape and failing to show the shape similarities between the target shape and other shape candidates in a retrieval list.
Thus we use Chamfer Distance (CD) and LFD L1~\cite{chen2003visual} to measure shape similarity given the GT and retrieved shapes.  We use $\text{CD}_k$ and $\text{LFD}_k$ to denote the average score over the top-K retrievals.
%CD makes an assumption that the 3D shapes for comparison are well-aligned and scaled.
For CD, we sample 4K points using farthest point sampling for each shape.
For LFD L1, we represent each 3D shape as its LFD features computed from a set of pre-rendered binary masks.
Assuming all 3D shapes are normalized and centered in the same canonical orientation, we render 200 views by placing cameras on the vertices of 10 randomly rotated dodecahedrons. We use average L1 distance over all views to measure shape similarity.

\subsection{View-dependent Metrics}
% We go beyond retrieval and reconstruction metrics by proposing a set of view-dependent metrics between objects in images and 3D shapes. 

In image queries, the object to be retrieved might be occluded by other objects or self-occluded, in which case matching the visible parts to the ground truth from a similar viewpoint is desirable. Hence, we propose a set of view-dependent metrics between objects in images and 3D shapes. These metrics can quantify retrieval performance for cases without corresponding 3D shape annotations.
To compute the view-dependent metrics, we render each retrieved 3D shape under the same viewpoint and pose as the object in the image.
% The motivation behind is that retrieved 3D shapes are supposed to match visual observation. Especially for objects with occlusion, it only requires to match visible area of objects in image. 
% We make all such pose information available as part of our \moos dataset.
% This pose information is fully accessible in MOOS.

\mypara{Mask IoU.}
%We render complete binary masks for retrieved 3D shapes using the pose and viewpoint of the corresponding object and image query.
% We leverage the object instance mask in the scene to mask out occluded image areas. The IoU is computed between the object instance mask and the occluded mask of the retrieved 3D shape.
The mask IoU is computed between the unoccluded (complete) rendered binary masks for the GT and retrieved 3D shape.  
%We render the complete binary masks using the pose and viewpoint for the GT object and the image query.
%It is also possible to mask out occluded image areas using the object instance mask.
\mypara{vLFD L1.}
Normally LFD is computed from a set of multiview binary masks to measure region and contour similarity. Here, we only consider a single-view LFD by concatenating region-based and contour-based descriptors for one mask.
The mask-to-mask LFD distance is defined as the $L1$ distance between LFDs of silhouettes of the ground truth and retrieved 3D shapes.
% \textbf{Normal IoU and Distance.} \todo{Probably move to supplement} Both mask IoU and vLFD only care about similarity of silhouette of object renderings without taking into account geometry details inside object rendering region. Therefore, we propose view-dependent normal distance to measure geometry similarity between target object and retrieved 3D shapes. We render normal maps for retrieved 3D shapes using the pose and viewpoint of the corresponding object and image query. Similar to the post-processing on mask renderings, we remove invisible parts of normal maps due to object occlusion. For each pair of GT normal and predicted normal, we compute their distance by averaging all pairwise angular distances of the normals the visible area. 
% % The distance is normalized to sum to 1. 
\mypara{LPIPS~\cite{zhang2018unreasonable}.}
The mask IoU and vLFD metrics only account for similarity of the object silhouette, ignoring content in the interior of the object.
To capture the perceptual similarity between the object as observed in the image and the retrieved shape, we feed resized patches of the query image and shape renders with occluded area masked out and cropped from the bounding box of the object mask into a pretrained VGG~\cite{simonyan2014very} model to compute LPIPS scores. Lower scores indicate better matches.

\begin{table*}
\centering
\resizebox{\linewidth}{!}
{
\begin{tabular}{@{}ll ccccccc ccccccc@{}}
\toprule
\multirow{2}{*}{Models} & \multicolumn{7}{c}{\textit{NoOcc} val set} & \multicolumn{7}{c}{\textit{Occ} val set} \\
\cmidrule(lr){2-8}\cmidrule(lr){9-15}
& $Acc_1$$\uparrow$ & $CatAcc$$\uparrow$ & CD$\downarrow$ & LFD$\downarrow$ & MIoU$\uparrow$ & vLFD$\downarrow$ & LPIPS$\downarrow$ & $Acc_1$$\uparrow$ & $CatAcc$$\uparrow$ & CD$\downarrow$ & LFD$\downarrow$ & MIoU$\uparrow$ & vLFD$\downarrow$ & LPIPS$\downarrow$ \\
\midrule
\textit{NoOcc}-CMIC & 84.1 & 99.3 & 0.631 & 0.430 & 0.961 & 0.137 & 0.159 & 48.3 & 85.9 & 2.090 & 1.391 & 0.735 & 1.023 & 0.224 \\
\textit{Occ}-CMIC & 86.1 & 99.5 & 0.552 & 0.372 & 0.966 & 0.125 & 0.157 & 81.6 & 98.4 & 0.755 & 0.499 & 0.938 & 0.260 & 0.149 \\
\textit{All}-CMIC & \textbf{87.8} & \textbf{99.7} & \textbf{0.485} & \textbf{0.330} & \textbf{0.972} & \textbf{0.108} & \textbf{0.153} & \textbf{82.6} & \textbf{98.6} & \textbf{0.696} & \textbf{0.469} & \textbf{0.941} & \textbf{0.251} & \textbf{0.147} \\
\bottomrule
\end{tabular}
}
\vspace{-8pt}
\caption{Cross-evaluation on the no occlusion (\textit{NoOcc}), occlusion (\textit{Occ}), and combined (\textit{All}) sets of \moos with seen objects during training. Training on all cases leads to the best performance in both occluded and occlusion-free image queries.}
\label{tab:cross_train_eval}
\end{table*}

\begin{figure}[t]
    \vspace{-12pt}
    \includegraphics[width=\linewidth]{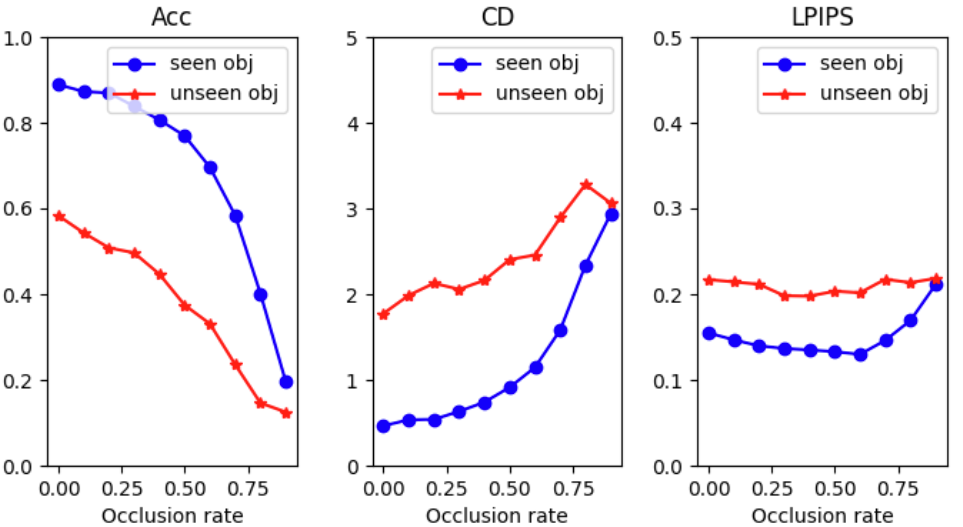}
    \vspace{-18pt}
    \caption{Plots of evaluation metrics vs. occlusion rates on MOOS seen (blue) and unseen (red) occluded objects.  Acc and CD decreases for higher occlusion rates while the LPIPS remains stable.}
    \label{fig:metrics_vc_occlusion}
\end{figure}

% \begin{table*}
% \centering
% \resizebox{\linewidth}{!}
% {
% \begin{tabular}{@{}ll ccc cccc cccccc@{}}
% \toprule
% \multirow{2}{*}{MOOS Set} & \multirow{2}{*}{Objects} & \multicolumn{3}{c}{Retrieval Accuracy} & \multicolumn{4}{c}{View-independent Metrics} & \multicolumn{6}{c}{View-dependent Metrics} \\
% \cmidrule(lr){3-5}\cmidrule(lr){6-9}\cmidrule(lr){10-15}
% & & $Acc_1$$\uparrow$ & $Acc_{5}$$\uparrow$ & \textit{CatAcc}$\uparrow$ & CD$\downarrow$ & $\text{CD}_5$$\downarrow$ & LFD$\downarrow$ & $\text{LFD}_5$$\downarrow$ & MIoU$\uparrow$ & $\text{MIoU}_5$$\uparrow$ & vLFD$\downarrow$ & $\text{vLFD}_5$$\downarrow$ & LPIPS$\downarrow$ & $\text{LPIPS}_5$$\downarrow$ \\
% \midrule
% \multirow{2}{*}{NoOcc} & seen & 87.8 & 99.4 & 99.7 & 0.4847 & 3.1431 & 0.3298 & 2.1578 & 0.9722 & 0.6534 & 0.1078 & 1.0645 & 0.1533 & 0.3097 \\
% & unseen & 58.9 & 88.7 & 94.9 & 1.7101 & 3.2584 & 1.1263 & 2.2293 & 0.8314 & 0.6200 & 0.5254 & 1.1950 & 0.2174 & 0.3217 \\
% \midrule
% \multirow{2}{*}{Occ} & seen & 82.6 & 96.6 & 98.6 & 0.6963 & 3.2039 & 0.4688 & 2.1766 & 0.9413 & 0.6609 & 0.2510 & 1.1679 & 0.1472 & 0.2742 \\
% & unseen & 49.6 & 79.4 & 93.4 & 2.0754 & 3.3499 & 1.3741 & 2.2771 & 0.7873 & 0.6291 & 0.7715 & 1.2982 & 0.2113 & 0.2863 \\
% \bottomrule
% \end{tabular}
% }
% \vspace{-8pt}
% \caption{Breakdown of evaluation on \moos for query objects seen during training and for unseen objects. The unseen object queries are much more challenging across all metrics.}
% \label{tab:test_on_unseen_object}
% \end{table*}

\begin{table*}
\centering
\resizebox{\linewidth}{!}
{
\begin{tabular}{@{}ll ccc cccc cccccc@{}}
\toprule
\multirow{2}{*}{MOOS Set} & \multirow{2}{*}{Objects} & \multicolumn{3}{c}{Retrieval Accuracy} & \multicolumn{4}{c}{View-independent Metrics} & \multicolumn{6}{c}{View-dependent Metrics} \\
\cmidrule(lr){3-5}\cmidrule(lr){6-9}\cmidrule(lr){10-15}
& & $Acc_1$$\uparrow$ & $Acc_{5}$$\uparrow$ & \textit{CatAcc}$\uparrow$ & CD$\downarrow$ & $\text{CD}_5$$\downarrow$ & LFD$\downarrow$ & $\text{LFD}_5$$\downarrow$ & MIoU$\uparrow$ & $\text{MIoU}_5$$\uparrow$ & vLFD$\downarrow$ & $\text{vLFD}_5$$\downarrow$ & LPIPS$\downarrow$ & $\text{LPIPS}_5$$\downarrow$ \\
\midrule
\multirow{2}{*}{NoOcc} & seen & 87.8 & 99.4 & 99.7 & 0.485 & 3.143 & 0.330 & 2.158 & 0.972 & 0.653 & 0.108 & 1.065 & 0.153 & 0.310 \\
& unseen & 58.9 & 88.7 & 94.9 & 1.710 & 3.258 & 1.126 & 2.229 & 0.831 & 0.620 & 0.525 & 1.195 & 0.217 & 0.322 \\
\midrule
\multirow{2}{*}{Occ} & seen & 82.6 & 96.6 & 98.6 & 0.696 & 3.204 & 0.469 & 2.177 & 0.941 & 0.661 & 0.251 & 1.168 & 0.147 & 0.274 \\
& unseen & 49.6 & 79.4 & 93.4 & 2.075 & 3.350 & 1.374 & 2.277 & 0.787 & 0.629 & 0.772 & 1.298 & 0.211 & 0.286 \\
\bottomrule
\end{tabular}
}
\vspace{-8pt}
\caption{Comparison on \moos for query objects seen during training vs for unseen objects. Unseen queries are more challenging.}
\label{tab:test_on_unseen_object}
\end{table*}

\begin{figure*}[t]
    \vspace{-12pt}
    \includegraphics[width=\linewidth]{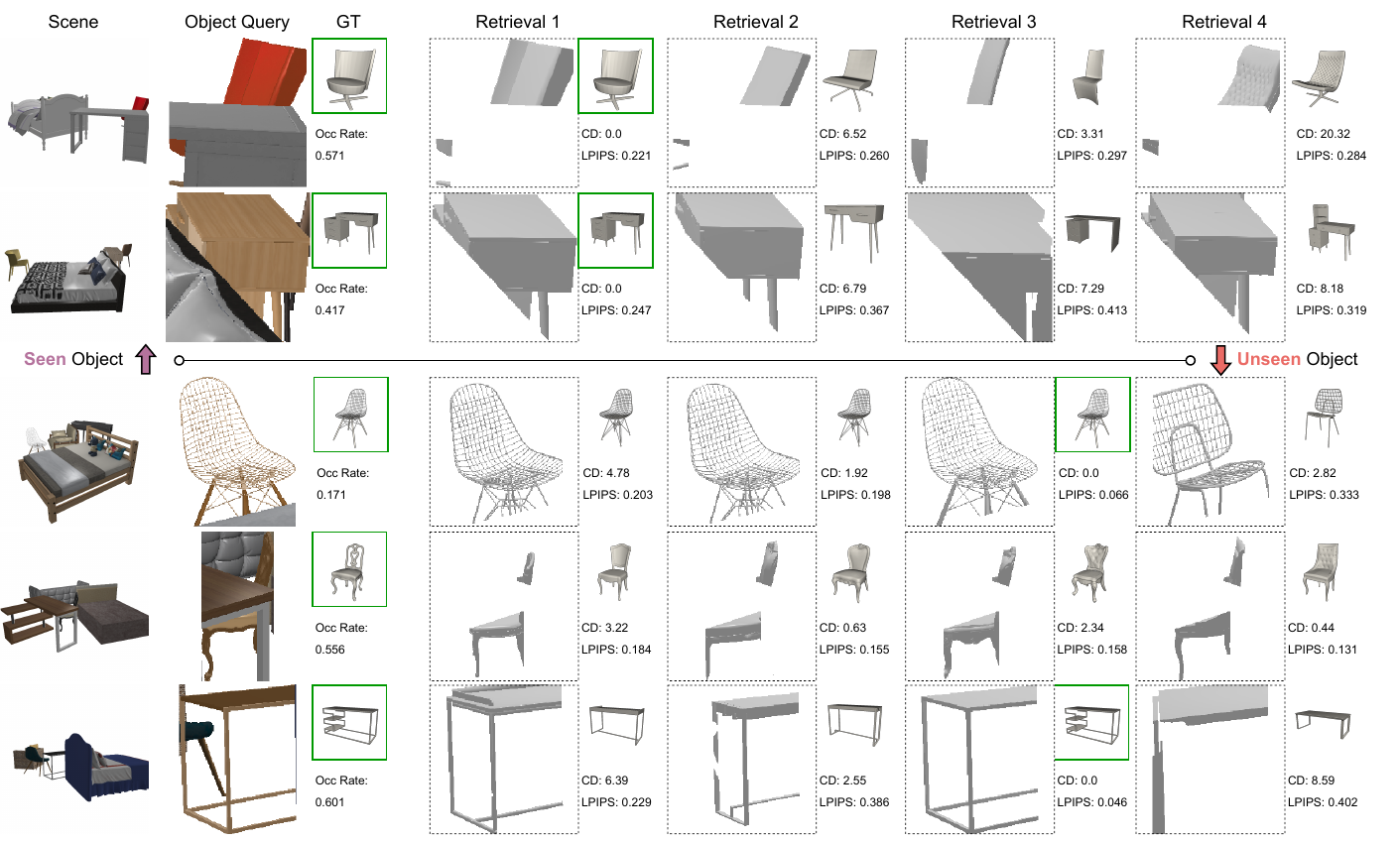}
    \vspace{-22pt}
    \caption{Results on MOOS seen (rows 1-2) and unseen (rows 3-5) objects.
    The target has an orange mask in the 2nd column . %The first two rows show objects seen during training. The last three rows show unseen objects. 
    For seen objects, we retrieve the matching object despite occlusions.
    For unseen objects, we see failures for objects with thin structures (row 3 and 5).}
    \label{fig:moos_seen_unseen}
\end{figure*}

\section{Results}
\label{sec:results}

\mypara{Retrieval for non-occluded and occluded objects}.
Prior work~\cite{fu2020hard, grabner2019location, lin2021single} trains and evaluates on images with non-occluded and non-truncated objects.
This is equivalent to our \textit{NoOcc} set.
% As object masks offered by Pix3D are silhouettes of annotated 3D shapes, they are clean and ignore potential occlusions which are far from masks obtained for real-life applications.
\Cref{tab:pix3d_easy_hard} shows that models trained only on such easy retrieval queries generalize poorly to hard cases with occlusions (\textit{Occ}).
In contrast, models trained on both unoccluded and occluded sets (\textit{All})  outperform models that
only have access to one of the sets. %only have access to either the \textit{NoOcc} or \textit{Occ} sets.
% Specifically, We train three CMIC models on three training sets of MOOS and separately, where the last one is the combination of \textit{NoOcc} and \textit{Occ}.
% We evaluate them on validation sets of \textit{NoOcc} and \textit{Occ} for seen objects during training.
\Cref{tab:cross_train_eval} shows that performance of \textit{NoOcc}-CMIC drops significantly when testing on occluded objects (${\sim}36\%$ drop on \textit{Acc}, ${\sim}1.46$ increase on CD).
\textit{Occ}-CMIC, performs not only much better than \textit{NoOcc}-CMIC on the \textit{Occ} set, but also achieves comparable results on the \textit{NoOcc} set, which is unsurprising since the occlusion rates of objects in \textit{Occ} span from 0 to 1 so there exist objects that are barely occluded.
The fact that \textit{All}-CMIC outperforms both \textit{NoOcc}-CMIC and \textit{Occ}-CMIC on all metrics provides evidence that single-view shape retrieval models trained to handle occluded and unoccluded objects together generalize better. 
\Cref{fig:metrics_vc_occlusion} shows how the metrics changes as the occlusion of the input images increases.
Note both Acc and CD degrade drastically with increasing occlusion rate, while LPIPS is relatively stable.
All remaining experiments are conducted with the \textit{All}-CMIC model.  %\Cref{fig:moos_seen_unseen} shows qualitative results. 

% \begin{table}
% \centering
% \resizebox{\linewidth}{!}
% {
% \begin{tabular}{@{}ll cc ccc@{}}
% \toprule
% \multirow{2}{*}{Val Set} & \multirow{2}{*}{Shape} & \multicolumn{2}{c}{View-independent Metrics} & \multicolumn{3}{c}{View-dependent Metrics} \\
% \cmidrule(lr){3-4}\cmidrule(lr){5-7}
% & & CD$\downarrow$ & LFD$\downarrow$ & MIoU$\uparrow$ & vLFD$\downarrow$ & LPIPS$\downarrow$ \\
% \midrule
% \multirow{2}{*}{\textit{NoOcc}} & 3D-FUTURE & 0.4847 & 0.3298 & 0.9722 & 0.1078 & 0.1533 \\
% & Scan2CAD & 5.4421 & 2.7352 & 0.4331 & 1.5744 & 0.4093 \\
% \midrule
% \multirow{2}{*}{\textit{Occ}} & 3D-FUTURE & 0.6963 & 0.4688 & 0.9413 & 0.2510 & 0.1472 \\
% & Scan2CAD & 5.4304 & 2.6899 & 0.4515 & 1.6950 & 0.3608 \\
% \bottomrule
% \end{tabular}
% }
% \vspace{-8pt}
% \caption{Evaluation on different shape database of ShapeNet 3D shapes from Scan2CAD on MOOS seen object images using the \textit{All}-CMIC model.  This regime is extremely challenging and as the shape database is different from the original annotated shape database, there is no ground-truth shape for measuring the accuracy.
% %using model trained with the \moos all set}.
% }
% \label{tab:test_on_unseen_shapes}
% \end{table}

\begin{table}
\centering
\resizebox{\linewidth}{!}
{
\begin{tabular}{@{}ll cc ccc@{}}
\toprule
\multirow{2}{*}{Val Set} & \multirow{2}{*}{Shape} & \multicolumn{2}{c}{View-independent Metrics} & \multicolumn{3}{c}{View-dependent Metrics} \\
\cmidrule(lr){3-4}\cmidrule(lr){5-7}
& & CD$\downarrow$ & LFD$\downarrow$ & MIoU$\uparrow$ & vLFD$\downarrow$ & LPIPS$\downarrow$ \\
\midrule
\multirow{2}{*}{\textit{NoOcc}} & 3D-FUTURE & 0.485 & 0.330 & 0.972 & 0.108 & 0.153 \\
& Scan2CAD & 5.442 & 2.735 & 0.433 & 1.574 & 0.409 \\
\midrule
\multirow{2}{*}{\textit{Occ}} & 3D-FUTURE & 0.69 & 0.469 & 0.941 & 0.251 & 0.147 \\
& Scan2CAD & 5.430 & 2.690 & 0.452 & 1.695 & 0.361 \\
\bottomrule
\end{tabular}
}
\vspace{-8pt}
\caption{Evaluation on different shape database of ShapeNet 3D shapes from Scan2CAD on MOOS seen object images using the \textit{All}-CMIC model.  This regime is extremely challenging and as the shape database is different from the original annotated shape database, there is no ground-truth shape for measuring the accuracy.
%using model trained with the \moos all set}.
}
\label{tab:test_on_unseen_shapes}
\end{table}

\mypara{Generalization to unseen object queries.}
Generalization to retrieval of 3D shapes for unseen object queries is crucial for deployment in practical applications.
% \citet{gkioxari2019mesh} creates two splits $S_1$ and $S_2$ for Pix3D to study the unseen object generalization issue, but \cite{lin2021single, grabner2019location, fu2020hard} do not utilize these splits.
Our \moos dataset contains unobserved object queries for both the \textit{NoOcc} and \textit{Occ} sets.
\Cref{tab:test_on_unseen_object} reports how the \textit{All}-CMIC model performs on the \textit{NoOcc} and \textit{Occ} val set for unseen objects.
Although the high $CatAcc$ numbers indicate shapes matching the correct category are retrieved, $Acc_1$ is noticeably lower by ${\sim}30\%$ than results for seen object queries. The view-independent metrics CD and LFD also increase by about 1.23 and 0.80 respectively for non-occluded objects. For view-dependent metrics, the small increase in LPIPS (${\sim}0.064$) shows that retrieved shapes are still perceptually similar to the ground truth shape under the same viewpoint.
\Cref{fig:moos_seen_unseen} shows examples for seen and unseen object queries.

\mypara{Generalization to similar shapes.}
We test generalization of our model trained on 3D-FUTURE shapes to similar 3D shapes by retrieving from a different shape database.  Instead of using 3D-FUTURE shapes as our retrieval database, we retrieve from 1,779 ShapeNet shapes used in Scan2CAD. 
%by replacing the 3D-FUTURE shapes used in training with 1,779 unseen shapes from ShapeNet used in Scan2CAD.
%\Cref{tab:test_on_unseen_shapes} reports the results.
Note that accuracy-based metrics are unavailable since there is no ground truth 3D shape.
Results (\Cref{tab:test_on_unseen_shapes}) show that this regime is quite challenging.
One important factor is that ShapeNet shapes are not as cleanly annotated as 3D-FUTURE.
In particular, incorrect shape dimensions strongly impact metrics by breaking the correct scale assumption.

\begin{table*}
\centering
\resizebox{\linewidth}{!}
{
\begin{tabular}{@{}l ccc cccc cccccc@{}}
\toprule
 Models & \multicolumn{3}{c}{Retrieval Accuracy} & \multicolumn{4}{c}{View-independent Metrics} & \multicolumn{6}{c}{View-dependent Metrics} \\
\cmidrule(lr){2-4}\cmidrule(lr){5-8}\cmidrule(lr){9-14}
& $Acc_1$$\uparrow$ & $Acc_{5}$$\uparrow$ & \textit{CatAcc}$\uparrow$ & CD$\downarrow$ & $\text{CD}_5$$\downarrow$ & LFD$\downarrow$ & $\text{LFD}_5$$\downarrow$ & MIoU$\uparrow$ & $\text{MIoU}_5$$\uparrow$ & vLFD$\downarrow$ & $\text{vLFD}_5$$\downarrow$ & LPIPS$\downarrow$ & $\text{LPIPS}_5$$\downarrow$ \\
\midrule
CMIC-Pix3D (500ep) & 47.6 & 66.2 & 94.1 & 0.761 & 1.429 & 1.399 & 2.353 & 0.765 & 0.607 & 0.780 & 1.341 & 0.311 & 0.335 \\
CMIC-MOOS-ft (5ep) & \textbf{55.1} & \textbf{80.3} & \textbf{95.5} & \textbf{0.608} & \textbf{1.319} & \textbf{1.234} & \textbf{2.279} & \textbf{0.808} & \textbf{0.625} & \textbf{0.650} & \textbf{1.273} & \textbf{0.299} & \textbf{0.330} \\
\midrule
CMIC-Scan2CAD (500ep) & 27.7 & 42.1 & 85.2 & 1.161 & 1.391 & 1.482 & 1.860 & 0.489 & 0.414 & \textbf{2.795} & 3.007 & 0.298 & 0.318 \\
CMIC-MOOS-ft (5ep) & \textbf{27.7} & \textbf{47.2} & \textbf{86.0} & \textbf{1.111} & \textbf{1.278} & \textbf{1.444} & \textbf{1.776} & \textbf{0.490} & \textbf{0.422} & 2.810 & \textbf{2.970} & \textbf{0.298} & \textbf{0.316} \\
\bottomrule
\end{tabular}
}
\vspace{-8pt}
\caption{Evaluation on Pix3D and Scan2CAD real image datasets. The pretrained-then-finetuned \moos models (CMIC-MOOS-ft) outperform baselines (CMIC-Pix3D and CMIC-Scan2CAD) trained on real datasets, demonstrating the benefit of our synthetic pretraining.}
\label{tab:transfer_from_syn_to_real}
\end{table*}

\begin{figure}[t]
    \vspace{-12pt}
    \includegraphics[width=\linewidth]{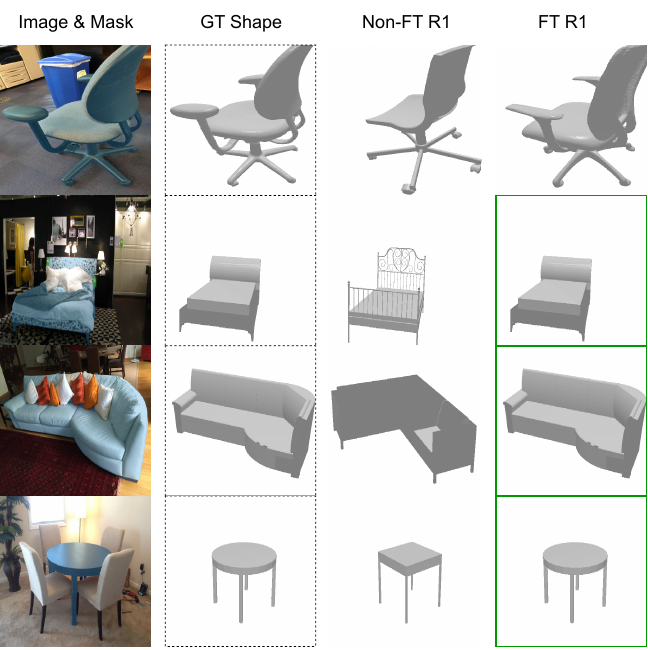}
    \vspace{-18pt}
    \caption{Top-1 retrievals on Pix3D unseen occluded objects with CMIC-Pix3D trained on Pix3D (Non-FT) and CMIC-MOOS-ft fine-tuned on Pix3D (FT). The object query is highlighted with a blue mask predicted using Mask2Former~\cite{cheng2022masked} from the RGB image. Green outlines are shapes matching the ground truth.}
    \label{fig:moos_ft_pix3d}
\end{figure}

\begin{figure}[t]
    \vspace{-16pt}
    \includegraphics[width=\linewidth]{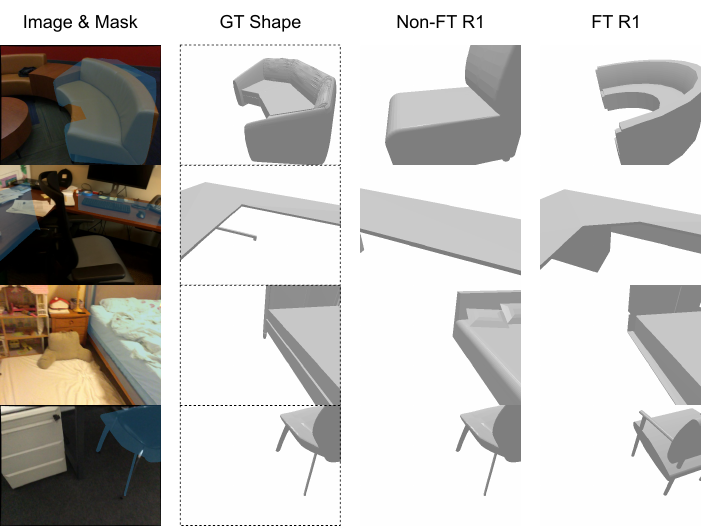}
    \vspace{-18pt}
    \caption{Top-1 retrievals on ScanNet using CMIC-Scan2CAD trained on Scan2CAD (Non-FT) and CMIC-MOOS-ft fine-tuned on Scan2CAD (FT). The query is highlighted with a blue mask projected from ground truth annotations. Here, the top@1 shapes do not match the ground truth but resemble the observed object.}
    \label{fig:moos_ft_scan2cad}
\end{figure}

\begin{table}
\centering
\resizebox{\linewidth}{!}
{
\begin{tabular}{@{}ll rrrr@{}}
\toprule
\multirow{2}{*}{Models} & \multirow{2}{*}{objects} & \multicolumn{2}{c}{Easy set} & \multicolumn{2}{c}{Hard set} \\
\cmidrule(lr){3-4}\cmidrule(lr){5-6}
& & $Acc_1\uparrow$ & $Acc_5\uparrow$ & $Acc_1\uparrow$ & $Acc_5\uparrow$  \\
\midrule
\multirow{2}{*}{CMIC-Pix3D} & seen & 74.9 & 89.2 & 47.2 & 73.5 \\
& unseen & 0.3 & 7.6 & 0.0 & 17.4  \\
\multirow{2}{*}{CMIC-MOOS-ft} & seen & 72.7 & \textbf{93.1} & \textbf{64.8} & \textbf{89.3} \\
& unseen & \textbf{34.3} & \textbf{62.8} & \textbf{24.2} & \textbf{62.4} \\
\bottomrule
\end{tabular}
}
\vspace{-8pt}
\caption{
Performance breakdown of CMIC models on Pix3D ``Easy set'' and ``Hard set''. Note the significant improvements of the MOOS pretrained model on unobserved objects.
}
\label{tab:pix3d_easy_hard_results}
\vspace{-18pt}
\end{table}

\mypara{Transfer from synthetic to real datasets.}
% The difficulty of collecting large-scale data of image-shape pairs exposes an important question how learning-based methods pretrained on synthetic data perform on realistic data.
We demonstrate transfer of CMIC models pretrained on \moos to two datasets with real images, Pix3D and Scan2CAD.
In \Cref{tab:transfer_from_syn_to_real}, we compare with CMIC-Pix3D and CMIC-Scan2CAD which are trained directly on the corresponding datasets for 500 epochs using the same hyperparameters.
CMIC-Pix3D uses predicted object segmentations from Mask2Former~\cite{cheng2022masked} instead of the ground truth shape mask. %since it is the rendering of shape silhouette that ignores occlusions on the object.
For CMIC-Scan2CAD, we use the object mask from the dataset which may reflect occlusions as multiple objects are present in scene.
CMIC-MOOS-ft is first pretrained on the \textit{All} set of \moos and then fine-tuned for 5 epochs on the respective real-world image dataset.
On Pix3D, CMIC-MOOS-ft significantly surpasses CMIC-Pix3D with 7.5\% higher accuracy, 0.153 lower CD, and 0.0122 lower LPIPS. It has better performance on objects unobserved or containing occlusions (see \Cref{tab:pix3d_easy_hard_results}).
On Scan2CAD, the pretrain-then-finetune strategy achieves competitive performance to CMIC-Scan2CAD.
Although retrieval accuracy is the same, better performance on view-independent and view-dependent metrics except vLFD implies that CMIC-MOOS-ft retrieves 3D shapes that are more similar to the desired 3D shape in terms of geometry.
\Cref{fig:moos_ft_pix3d,fig:moos_ft_scan2cad} shows qualitative examples.
%See supplement for more details on transfer performance under few-shot fine-tuning.

% \begin{table}
% \centering
% \resizebox{\linewidth}{!}
% {
% \begin{tabular}{@{}l cccccc@{}}
% \toprule
% %  & \multicolumn{3}{c}{Retrieval Accuracy} & \multicolumn{4}{c}{Shape-wise Metrics} & \multicolumn{6}{c}{View-dependent Metrics} \\
% % \cmidrule(lr){2-4}\cmidrule(lr){5-8}\cmidrule(lr){9-14}
% Models & $Acc_1$$\uparrow$ & CD$\downarrow$ & LFD$\downarrow$ & MIoU$\uparrow$ & vLFD$\downarrow$ & LPIPS$\downarrow$ \\
% \midrule
% CMIC-Pix3D & 47.6 & 0.7614 & 1.3999 & 0.7642 & 0.7852 & 0.3103  \\
% \midrule
% 0-shot inference & 20.9 & 1.5101 & 2.1350 & 0.6236 & 1.2858 & 0.3601 \\
% 2\%-shot-ft & 29.1 & 1.2756 & 1.9188 & 0.6689 & 1.1191 & 0.3441  \\
% 10\%-shot-ft & 44.1 & 0.9042 & 1.5173 & 0.7508 & 0.8422 & 0.3118  \\
% 50\%-shot-ft & 50.1 & 0.7305 & 1.3643 & 0.7838 & 0.7197 & 0.2992  \\
% 100\%-shot-ft & 55.1 & 0.6134 & 1.2377 & 0.8055 & 0.6542 & 0.2918  \\
% \bottomrule
% \end{tabular}
% }
% \vspace{-8pt}
% \caption{Few-shot MOOS pretrained CMIC fine-tuned on Pix3D.  We show that we can get reasonable performance even with just fine-tuning on a small fraction of the Pix3D data.}
% \label{tab:few_shot_pix3d}
% \end{table}

\begin{table}
\centering
\resizebox{\linewidth}{!}
{
\begin{tabular}{@{}l cccccc@{}}
\toprule
%  & \multicolumn{3}{c}{Retrieval Accuracy} & \multicolumn{4}{c}{Shape-wise Metrics} & \multicolumn{6}{c}{View-dependent Metrics} \\
% \cmidrule(lr){2-4}\cmidrule(lr){5-8}\cmidrule(lr){9-14}
Models & $Acc_1$$\uparrow$ & CD$\downarrow$ & LFD$\downarrow$ & MIoU$\uparrow$ & vLFD$\downarrow$ & LPIPS$\downarrow$ \\
\midrule
CMIC-Pix3D & 47.6 & 0.761 & 1.400 & 0.764 & 0.785 & 0.310  \\
\midrule
0-shot inference & 20.9 & 1.510 & 2.135 & 0.624 & 1.286 & 0.360 \\
2\%-shot-ft & 29.1 & 1.276 & 1.919 & 0.669 & 1.119 & 0.344  \\
10\%-shot-ft & 44.1 & 0.904 & 1.517 & 0.751 & 0.842 & 0.312  \\
50\%-shot-ft & 50.1 & 0.731 & 1.364 & 0.784 & 0.720 & 0.299  \\
100\%-shot-ft & 55.1 & 0.613 & 1.238 & 0.806 & 0.654 & 0.292  \\
\bottomrule
\end{tabular}
}
\vspace{-8pt}
\caption{Few-shot MOOS pretrained CMIC fine-tuned on Pix3D.
Even fine-tuning on fractions of Pix3D gives good performance.}
\label{tab:few_shot_pix3d}
\end{table}
\mypara{Few-shot fine-tuning.}
We conduct few-shot fine-tuning on Pix3D (see \Cref{tab:few_shot_pix3d}) where $n$\%-shot-ft indicates a CMIC model pretrained on \moos and finetuned on $n$\% Pix3D. We show that training on MOOS data enables fine-tuning performant models on Pix3D data with considerably less data. The relatively poor results of zero-shot inference on Pix3D reflect the domain shift from synthetic to real images. When finetuning on 50\% Pix3D data, we still outperform CMIC-Pix3D. Even with only 10\% Pix3D data, performance is comparable to the whole training data. We conclude that synthetic data is useful for training generalizable models and the effort spent on data collection and annotation for real image-shape pairs can be significantly reduced.

% \subsection{Image Completion Guided Shape Retrieval}
% Object masks in reality are never as perfect as ground truth. \todo{We can probably evaluate results with view-dependent metrics}

% \subsection{Single-view Scene Generation}

\section{Conclusion}
\label{sec:conclusion}

% \textbf{Limitations}
% \todo{Metric limitation, LPIPS}
% \todo{generalize to unseen categories}

We studied the generalization of single-view 3D shape retrieval to occlusions and unseen objects.
We standardized two real image datasets and presented a synthetic dataset generation pipeline that allowed us to systematically evaluate the performance of shape retrieval for inputs with occlusions and with 3D shapes or objects unseen during training.
We show that training on synthetic data with occlusions helps significantly improve performance.
Though results are promising, the task remains challenging and in particular generalization to unseen categories of objects is an open question for future work.
We hope our work enables more rigorous evaluation of single-view 3D shape retrieval in practical settings.

% \todo{Add acknowledgement}
\mypara{Acknowledgments.}
This work was funded in part by a CIFAR AI Chair, a Canada Research Chair, NSERC Discovery Grant, NSF award \#2016532, and enabled by support from \href{https://www.westgrid.ca/}{WestGrid} and \href{https://www.computecanada.ca/}{Compute Canada}. Daniel Ritchie is an advisor to Geopipe and owns equity in the company. Geopipe is a start-up that is developing 3D technology to build immersive virtual copies of the real world with applications in various fields, including games and architecture. We thank Weijie Lin for help with initial development of the metrics code, and Sonia Raychaudhuri, Yongsen Mao, Sanjay Haresh and Yiming Zhang for feedback on paper drafts.

{\small
\bibliographystyle{plainnat}
\setlength{\bibsep}{0pt}
\bibliography{main}
}

\ifarxiv 
% \ifreview
\clearpage \appendix
\label{sec:appendix}

In this supplement, we provide more details about the CMIC model (\Cref{sec:supp-model-details}), and the generation process (\Cref{sec:supp-layout}) and statistics (\Cref{sec:supp-data-statistics}) for our synthetic \moos dataset.
We also provide a detailed analysis of metrics for shape retrieval (\Cref{sec:supp-metrics}), and additional results (\Cref{sec:supp-results}).

\section{CMIC Model Details}
\label{sec:supp-model-details}

We present an overview of the CMIC~\cite{lin2021single} architecture in \Cref{fig:cmic-arch}. Given a RGB image that may contain multiple objects and a binary mask indicating the object of interest obtained from either ground-truth or prediction, the object RGB query and mask are cropped from the input image and mask using 2D bounding box computed from the input mask. All image inputs, including object query, object mask and 3D shape multiview renderings, are resized to 224x224
for passing to the image encoders.  For the image encoders, we use  ResNet~\cite{he2016deep}-based encoders pretrained on ImageNet~\cite{krizhevsky2012imagenet}, R50 for the object query and R18 for the rendered shape images. We adapt the encoders for our task by modifying the first convolution layer to take inputs with 4-channel (RGB+mask) for the object query (R50) and 1-channel (greyscale) for the shape multiview renderings (R18). The output dimensions of the last FC layers of the encoders are also modified to obtain embeddings of dimension 128.
Given query features $\{f^i\}_{i=1}^B$ and query-attended shape features $\{f^{S_k}_i\}_{i=1}^B$, we define the similarity function $D(x,y)$ with a temperature hyperparameter $\tau$ as follows:
$$
    D(x, y):=e^{\frac{1}{\tau}\left(\frac{x}{\|x\|}\right)^{T}\left(\frac{y}{\|y\|}\right)}
$$

\begin{figure}
    \includegraphics[width=\linewidth]{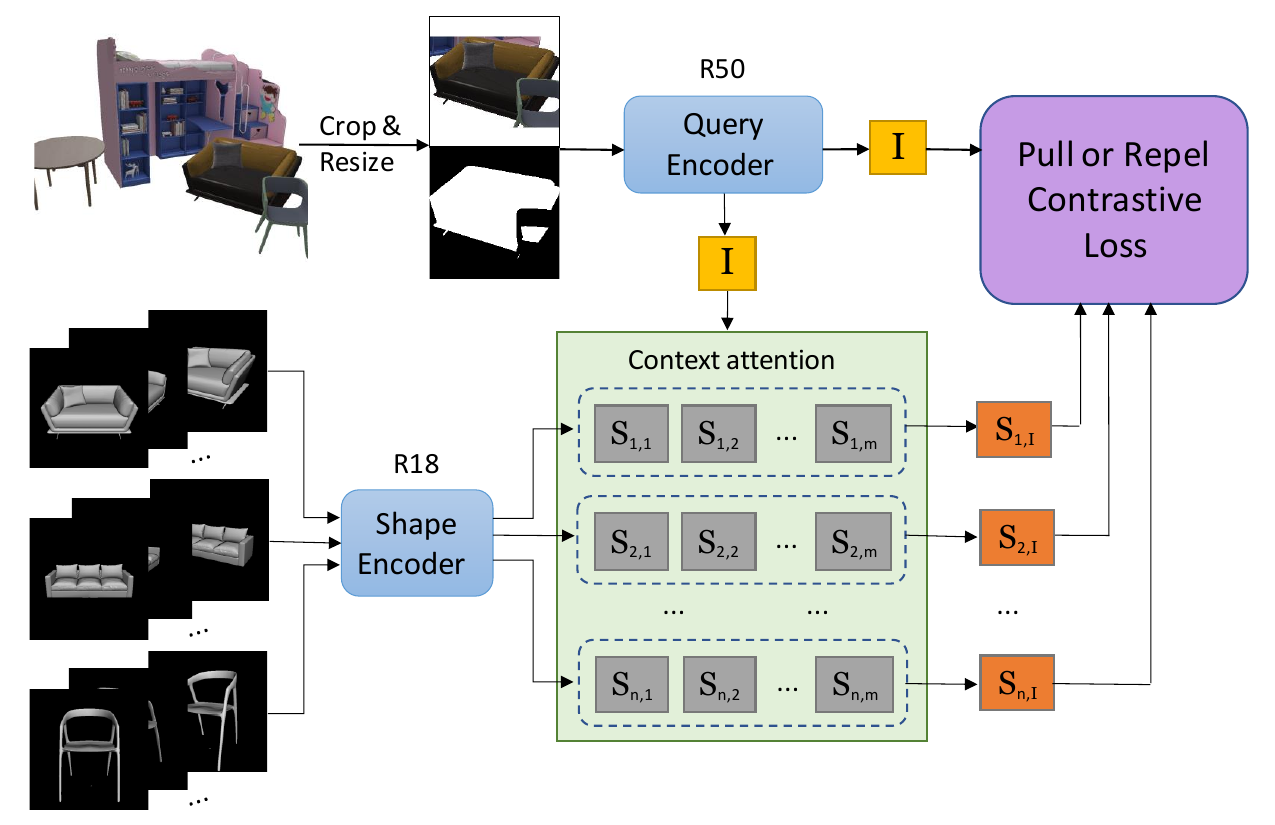}
    \caption{Overview of the CMIC~\cite{lin2021single} architecture that we implement in this work as the basis of our experiments. Shapes are encoded by rendering multiview images in gray-scale and passing them to a R-18 model.  Crops of object queries are extracted from the input image using the mask and encoded using R-50 model. 
 A mix of instance-level and class-level contrastive loss is used to train the CMIC model.}
    \label{fig:cmic-arch}
\end{figure}

\section{Multi-Object Occlusion Scene}
\label{sec:supp-dataset}

\subsection{Layout generation}
\label{sec:supp-layout}

The layout generation for a multi-object occlusion scene is implemented by iteratively inserting newly sampled 3D shapes into the existing layout to make sure its 2D bounding box does not intersect with any 2D bounding boxes of previously selected 3D shapes. Each scene is composed of 4 3D-FUTURE~\cite{fu20213d} objects from 4 categories (chair, bed, table and sofa) respectively. The goal is to form a scene where objects are close to each other to observe occlusion, but they don’t overlap.
We first randomly sample 4 objects from 4 categories.
% Scale each object to its pre-normalized dimensions. Set the starting position to the origin (0,0)
For each 3D shape $S_i$ remaining to be placed, we rotate it by a random angle around the up axis. The initial position $\pmb{p}_i^0$ is initialized as the origin if no shapes are placed before, otherwise set to the average position of all placed shapes. Then a unit vector $\pmb{v}_i$ is randomly sampled as a moving direction. A base distance scalar moving distance $d_i^0$ is the sum of short sides of all placed objects. The position of the new shape is then $d_i \cdot \pmb{v}_i + \pmb{p}_i^0$.
If there are any intersections between 2D bounding boxes of placed objects and the new object, we iterative increase $d_i$ by 0.05 until no intersections are observed after $N$ times. The final shape position is calculated as $(d_i^0 + 0.05 \times N) \cdot \pmb{v}_i + \pmb{p}_i^0$ (see the red dashed arrows in the top-down view layouts in the main paper). The vertical position of the shape is set so that the shape is positioned on the floor plane. We repeat the procedure  until all shapes are placed into the scene. Note that we only consider intersections among 2D bounding boxes from the top-down view as a simplification instead of using more expensive physics-based collision checks.
% Set the starting position to the mean of positions of all placed objects.
% Repeat step4 ~ step7 until all objects are placed.

\subsection{MOOS Statistics}
\label{sec:supp-data-statistics}

\begin{table}
\resizebox{\linewidth}{!}
{
\begin{tabular}{@{} ll rrrrr@{}}
\toprule
\multirow{2}{*}{Dataset} & \multirow{2}{*}{Split} & \multirow{2}{*}{Train Set} & \multicolumn{2}{c}{Val Set} & \multicolumn{2}{c}{Test Set} \\
\cmidrule(lr){4-5}\cmidrule(lr){6-7}
& & & Seen & Unseen & Seen & Unseen \\
\midrule
\multirow{2}{*}{Pix3D~\cite{sun2018pix3d}} & Easy & 2,998 (143) & 1,065 & 1,055 (179) & - & - \\
& Hard & 2,451 (143) & 903 & 178 (62) & - & - \\
% Scan2CAD~\cite{avetisyan2019scan2cad} & & 39,239 & & 11,671 \\
Scan2CAD~\cite{avetisyan2019scan2cad} & & 39K (1548) & \multicolumn{2}{c}{11K (560)} & - & - \\
\midrule
\multirow{2}{*}{MOOS} & NoOcc & 85K (5K) & 10K & 6K (618) & 12K & 6K (618) \\
& Occ & 249K (5K) & 31K & 17K (618) & 35K & 17K (618) \\
& All & 334K (5K) & 41K & 23K (618) & 47K & 23K (618) \\
\bottomrule
\end{tabular}
}
\vspace{-8pt}
\caption{Statistics of the number of object queries in the splits for the different datasets. Numbers in parentheses represent the number of unique 3D shapes.  For the seen split, the correct shape is found in the training set.}
\label{tab:data_splits_stats}
\end{table}

% \begin{table}
% \resizebox{\linewidth}{!}
% {
% \begin{tabular}{@{} l rrrrr@{}}
% \toprule
%  & Chair & Bed & Table & Sofa & All \\
% \midrule
% \# unique shapes & 1,639 & 1,073 & 1,038 & 2,459 & 6,209 \\
% \# objects w/ occlusion & 66,542 & 67,679 &  66,409 & 68,465 & 269,095 \\
% \# objects w/o occlusion & 48,106 &  52,242& 50,832 & 49,793 & 200,973 \\
% \# invisible objects & 4,656 &  76 & 2,676 &  1,704 & 9,112 \\
% \bottomrule
% \end{tabular}
% }
% \caption{The statistics of objects in MOOS for each category.}
% \label{tab:moos_object_stats}
% \end{table}

\begin{table}
\resizebox{\linewidth}{!}
{
\begin{tabular}{@{} l rrrrr@{}}
\toprule
 & Chair & Bed & Table & Sofa & All \\
\midrule
\# unique shapes & 1,639 & 1,073 & 1,038 & 2,457 & 6,207 \\
\# objects w/ occlusion & 78,745 & 93,877 & 86,807 & 90,580 & 350,009 \\
\# objects w/o occlusion & 35,752 & 26,033 & 30,335 & 27,621 & 119,741 \\
\# invisible objects & 5,503 & 90 & 2,858 & 1,799 & 10,250 \\
\bottomrule
\end{tabular}
}
\vspace{-8pt}
\caption{Statistics of rendered objects in MOOS generation for each category.  We report the number of number of unique shapes as well as number of object queries with and without occlusions.  In some generated scenes, some objects are completely occluded or truncated.  We do not includes those in our dataset of training or evaluation samples.}
\label{tab:moos_object_stats}
\end{table}

\begin{figure}
    \includegraphics[width=\linewidth,trim={0 0.2cm 0 0.2cm},clip]{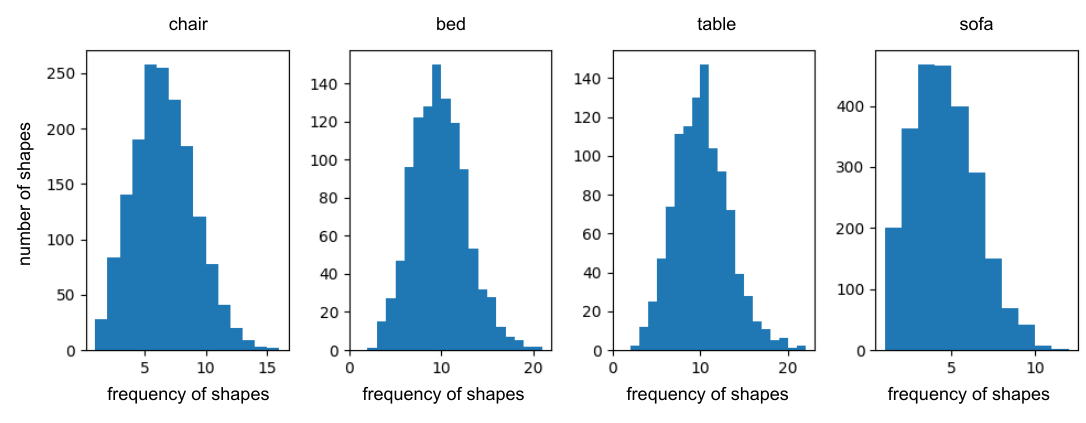}
    \caption{The number of unique shapes vs. the frequency of shapes for each category in \moos.}
    \label{fig:moos_obj_freq}
\end{figure}

\begin{figure}
    \includegraphics[width=\linewidth,trim={0 0.2cm 0 0.2cm},clip]{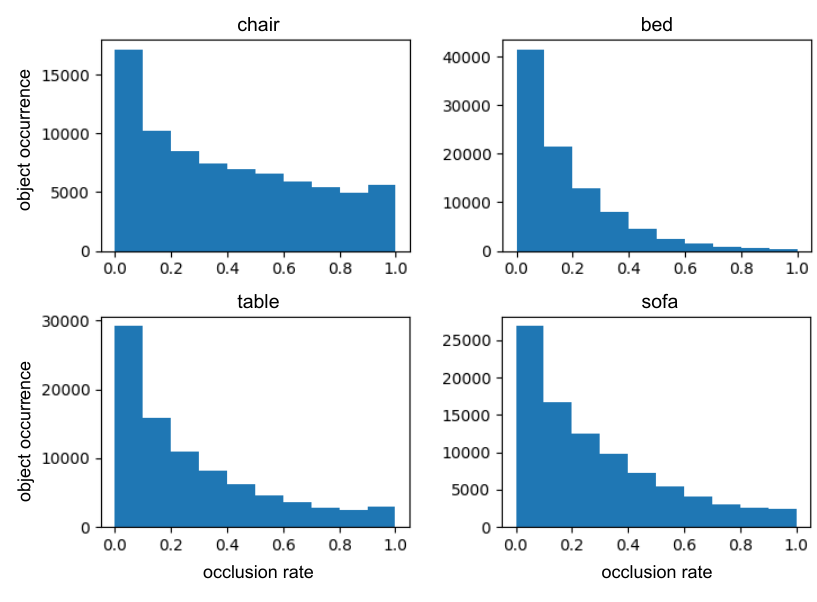}
    \caption{The count of object occurrences vs. occlusion rates for each category in \moos.}
    \label{fig:moos_obj_occlusion}
\end{figure}

In \Cref{tab:data_splits_stats}, we provide statistics for the train and val split we create for MOOS.  As part of of our work, we also provide standardized splits for image-to-shape retrieval with Pix3D~\cite{sun2018pix3d} with splits that includes both seen and unseen shapes in the validation set.  For  Scan2CAD~\cite{dai2017scannet}, the validation set consists of only unseen shapes.  Compared to the other two datasets, MOOS provides a large set of rendered images with and without occlusion for training.
In the NoOcc set, among all object queries we have 85K for training, 16K for validation, and 18K for test.
In the Occ set, among all object queries we have 249K for training, 48K for validation, and 52K for test.
For the 618 3D shapes that we set aside for exploring generalization to unseen objects, we have 23K object queries for validation and 23K for test.

We list detailed statistics of the objects in \moos in \Cref{tab:moos_object_stats} where for each category we count the number of unique shapes, the number of object instances in all scene renderings with/without some occlusions and the number of invisible objects due to complete occlusion or truncation.
In \Cref{fig:moos_obj_freq}, we plot a histogram of the number of unique shapes vs. the frequency of shapes for each category in \moos.
In \Cref{fig:moos_obj_occlusion}, we plot a histogram of the number of object occurrences vs. occlusion rates for each category in \moos.

\section{Metrics Analysis}
\label{sec:supp-metrics}

A variety of different metrics for comparing shape similarity has been proposed and used for shape retrieval.  Here we describe the analysis that we conduct to determine which metrics are more appropriate for single-view shape retrieval.  We use the metrics we find to work well in our main paper. To conduct our analysis, we consider a target shape query, and rank all 3D shapes in the database using a given similarity metric to determine the usability and stability of the metric for retrieval.

We categorize metric candidates into three groups: (1) Point-cloud based reconstruction metrics including CD, NC and $\text{F1}^t$; (2) Shape2shape metrics including voxel IoU, neural shape descriptor and LFD~\cite{chen2003visual} and (3) View-dependent metrics including mask IoU, vLFD, normal IoU, normal L2 and LPIPS~\cite{zhang2018unreasonable}.
Since our practical goal is to select metrics appropriate for the single-view shape retrieval task we perform a qualitative analysis over several hundred query examples.  
A rigorous study of metric design for single-view shape retrieval is an interesting direction for future work.

Based on our analysis we select the reconstruction metric CD, LFD as a view-based shape2shape metric, and several view-dependent metrics (mask IOU, vLFD, and LPIPS).  In the following sections, we provide more details on how we selected these metrics.

\subsection{Reconstruction Metrics}
\label{sec:supp-metrics-recons}

\begin{figure}
    \includegraphics[width=\linewidth,trim={0 0.2cm 0 0.2cm},clip]{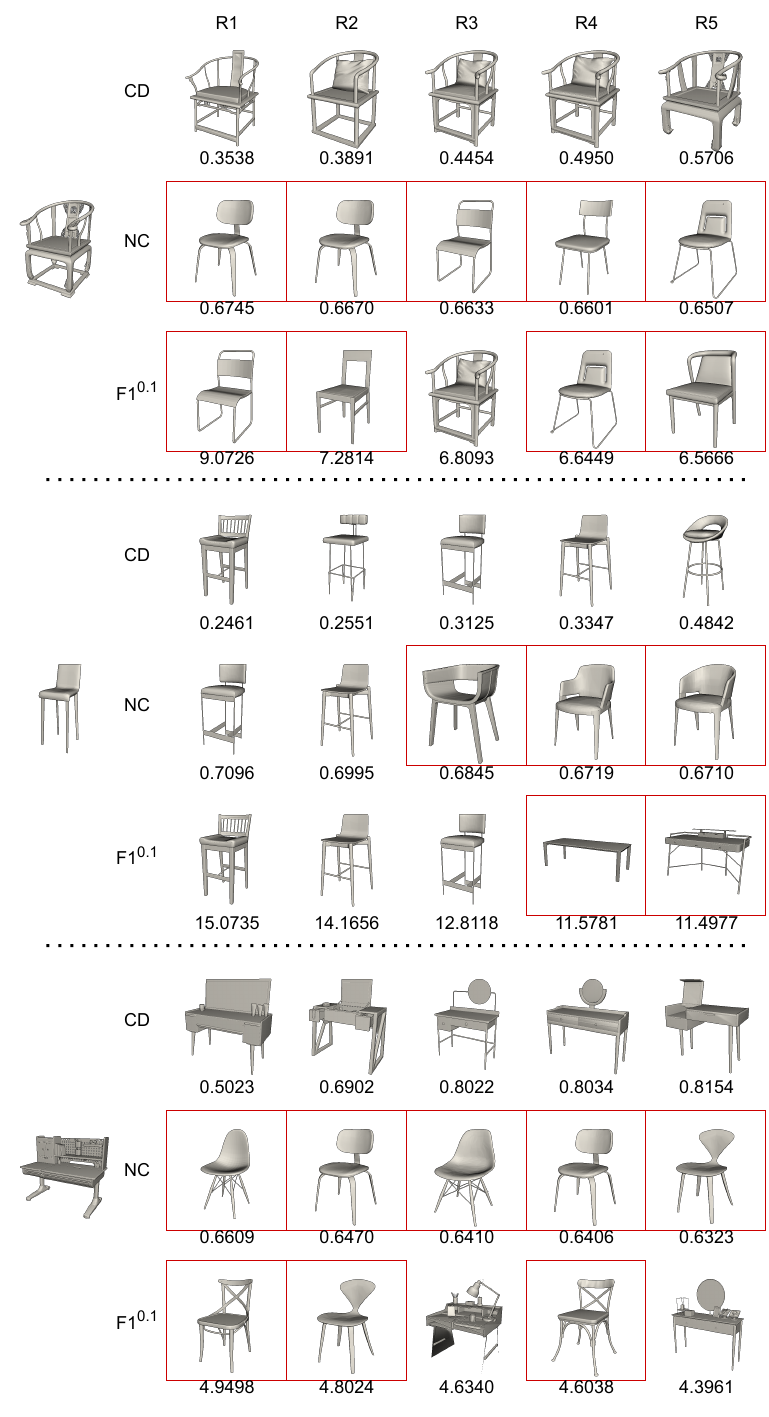}
    \caption{Comparison of shape rankings under three \textbf{point-cloud based reconstruction metrics} computed from 2K sampled points using face area weighted sampling. We show top-5 rankings excluding the ground-truth. The red box indicates a shape is noticeably different from the query shape due to distinguishable geometry or even wrong category. For CD, lower scores indicate shapes are more similar, while for NC and F1$^{0.1}$, higher scores indicate more similar shapes.  We find that CD gives more accurate retrieval results compared to NC and F1$^{0.1}$.}
    \label{fig:pc_metric_comparison}
\end{figure}

Reconstruction metrics computed from sampled point clouds are widely used to measure the quality of reconstructed or generated shapes against a target shape~\cite{wang2018pixel2mesh,gkioxari2019mesh}. These metrics assume that the 3D shapes being compared are well aligned and scaled.
Several prior work~\cite{sun2018pix3d,kuo2020mask2cad, kuo2021patch2cad, lin2021single} has opted to use point-wise reconstruction metrics to evaluate the single-view shape retrieval task. 

We consider three popular reconstruction metrics: Chamfer distance (CD), Normal consistency (NC), and point-wise F1 at distance threshold $t$ (F1$^t$) for measuring the similarity between two point clouds.
Given two point clouds \(P\), \(Q\) with positions and normals, let \(\Lambda_{P, Q}=\left\{\left(p, \arg \min _{q}\|p-q\|\right) | p \in P, q \in Q\right\}\) denote the set of pairs \((p,q)\) where \(q\) from \(Q\) is the nearest neighbor of p and \(n_{p}\) denote the unit normal vector at point \(p\). 

The Chamfer distance (CD) and Normal consistency (NC) measures the similarity of two point clouds based on either distance of points (CD) or similarity of unit normals (NC).  Formally, \textbf{CD} and \textbf{NC} are defined as:
\begin{align*}
\resizebox{1.0\hsize}{!}{$\mathrm{CD}(P, Q)=\frac{1}{2}\left(\frac{\sum_{(p, q) \in \Lambda_{P, Q}}d(p,q)}{|P|} + \frac{\sum_{(q, p) \in \Lambda_{Q, P}}d(p,q)}{|Q|}\right)$} \\
\resizebox{1.0\hsize}{!}{$\mathrm{NC}(P, Q)=\frac{1}{2}\left(\frac{\sum_{(p, q) \in \Lambda_{P, Q}}n_{p} \cdot n_{q}}{|P|} + \frac{\sum_{(q, p) \in \Lambda_{Q, P}}n_{q} \cdot n_{p}}{|Q|}\right)$}
\end{align*}
The $\textbf{F1}^t$ score measures the percentage of points that are accurately reconstructed by taking into account both precision (how close reconstructed points are to ground truth points) and completeness (the percentage of ground truth points that are covered) for a distance threshold $t$ that controls the strictness of score.
It is robust to the geometric layout of outliers~\cite{tatarchenko2019single} but does not have a scale-invariance property.
We set $t=0.1$ for our analysis.

% We choose point-wise reconstruction F1 over chamfer distance (CD) because CD can be significantly perturbed depending on how far the outlier points are from the ground truth shape, while F1 score is robust to the geometric layout of outliers~\cite{tatarchenko2019single}. 
% $\mathrm{F}1^{t}$ is conditioned on a threshold $t$ that indicates the strictness of score and lower threshold reflect the ability to reconstruct under higher criterion. A target point is considered to be successfully reconstructed by another point if their $L2$ distance is less than $t$.
% We compute \(\mathrm{F} 1^{0.1}\), \(\mathrm{F} 1^{0.3}\) and \(\mathrm{F} 1^{0.5}\) to confirm our method works well at different levels. 
% \todo{cite} also adopt $\text{AP}^{\text{mesh}}$ proposed by \todo{cite} that computes the average area under the precision-recall curve of $\mathrm{F} 1^{0.3}$ at IoU 0.5 (AP50) where a retrieved shape is regarded as a true positive shape if its $\mathrm{F} 1^{0.3}$ score is greater than 0.5. $\text{AP}^{\text{mesh}}$ is calculated for each category as well.

\begin{table}
\centering
% \resizebox{\linewidth}{!}
{
\begin{tabular}{@{}l cc@{}}
\toprule
Score Variance$\downarrow$ & FAS & FPS  \\
\midrule
CD & 5.11e-03 & \textbf{2.25e-03}   \\
NC & \textbf{6.49e-05} & 6.95e-05  \\
$\text{F1}^{0.1}$ & \textbf{54.89} & 91.14  \\
\bottomrule
\end{tabular}
}
\caption{Metric score variance over different number of points.  We compute the variance of the score for shape-to-shape pairs with different samplings of number of points (1K, 2K, 4K, and 10K), and compare the variance for face area-weighted sampling (FAS) vs farthest point sampling (FPS).  We find that $\text{F1}^{0.1}$ has relatively high variance (note that we report NC from 0 to 1, while we report $\text{F1}^{0.1}$ from 0 to 100), and that the combination of CD with FPS gives the least variance. }
\label{tab:pc_score_variance}
\end{table}

\begin{table}
\centering
\resizebox{\linewidth}{!}
{
\begin{tabular}{@{}ll ccc ccc@{}}
\toprule
\multirow{2}{*}{Metrics} & \multirow{2}{*}{\#points} & \multicolumn{3}{c}{FAS} & \multicolumn{3}{c}{FPS} \\
\cmidrule(lr){3-5}\cmidrule(lr){6-8}
& & mMS$\downarrow$ & mRD$\downarrow$ & RBO$\uparrow$ & mMS$\downarrow$ & mRD$\downarrow$ & RBO$\uparrow$ \\
\midrule
\multirow{3}{*}{CD} & 1K & 13.74 & 2.90 & 0.90 & 12.59 & 2.50 & 0.92 \\
& 2K & 11.62 & 2.12 & 0.94 & 10.97 & 1.96 & 0.95 \\
& 4K & 9.91 & 1.65 & 0.96 & \textbf{8.61} & \textbf{1.51} & \textbf{0.97} \\
\midrule
\multirow{3}{*}{NC} & 1K & 16.60 & 7.62 & 0.78 & 16.53 & 8.24 & 0.78 \\
& 2K & 15.79 & 5.06 & 0.83 & 15.73 & 5.40 & 0.83 \\
& 4K & 14.82 & 3.70 & 0.88 & 14.64 & 3.76 & 0.88\\
\midrule
\multirow{3}{*}{$\text{F1}^{0.1}$} & 1K & 17.54 & 26.81 & 0.64 & 17.69 & 33.05 & 0.60 \\
& 2K & 16.81 & 9.88 & 0.75 & 17.25 & 12.67 & 0.71 \\
& 4K & 15.59 & 4.95 & 0.84 & 16.13 & 6.25 & 0.81\\
\bottomrule
\end{tabular}
}
\caption{Stability of pointcloud-based reconstruction metrics for shape ranking with different number of sampled points and comparing face area weighted sampling (FAS) vs farthest point sampling (FPS). For measuring stability, we use the mean number of moved shape ranks (mMS), the mean rank distance (mRD) of moved shapes, and the rank biased overlap (RBO).  We find that CD with FPS gives the most stable rankings.}
\label{tab:pc_metrics_stability}
\end{table}

We compare the ranking quality among the three reconstruction metrics mentioned above computed from 2K sampled points using face area--weighted sampling (see examples in \Cref{fig:pc_metric_comparison}). Under the same sampling condition, NC and $\text{F1}^t$ tend to rank unrelated shapes higher compared to CD, which either present distinguishable geometric structure or have wrong categories. Note that the ranking score and result produced by F1 is subject to the threshold $t$ and the shape scale. With a larger threshold, $\text{F1}^{t}$ can rank shapes better. Based on our findings, we argue that CD is a simple and representative point cloud-based metric for shape retrieval that outputs reasonable shape ranking without the need to tweak hyperparameters. 

We also investigate the effect of the number of points and the sampling method on the shape ranking. For each shape, we sample 1K, 2K, 4K and 10K points using face area-weighted sampling (FAS) and farthest point sampling (FPS). We observe that FPS is better at capturing thin and small structure of the shape when the number of points is small. For CD, FPS results in smaller metric score variance over different number of points (see \Cref{tab:pc_score_variance}). We quantify the \emph{stability} of each metric under different number of points and sampling methods (see \Cref{tab:pc_metrics_stability}) using a set of stability metrics that we define. Given a base shape ranking computed from 10K points and a different sampling setting, we calculate the mean moved shapes (mMS) measuring how many shapes moved their ranks on average, the mean rank difference (mRD) over moved shapes,  and Rank Biased Overlap~\cite{webber2010similarity} (RBO) measuring the similarity between two ranked lists. CD with FPS sampling gives the most stable rankings over different number of points. Considering both computation efficiency and stability, we find that CD with 4K sampled points using farthest point sampling is an appropriate metric for retrieval task. We note that \citet{sun2018pix3d} has conducted a user-study and found that CD and EMD to correlate better with human judgement than voxel IoU, with CD having the best correlation.  One weakness of CD is that its score is potentially less interpretable than NC and $\text{F1}^{t}$.

\subsection{Shape2shape Metrics}
\label{sec:supp-metrics-shape}

\begin{figure}
    \includegraphics[width=\linewidth,trim={0 0.2cm 0 0.2cm},clip]{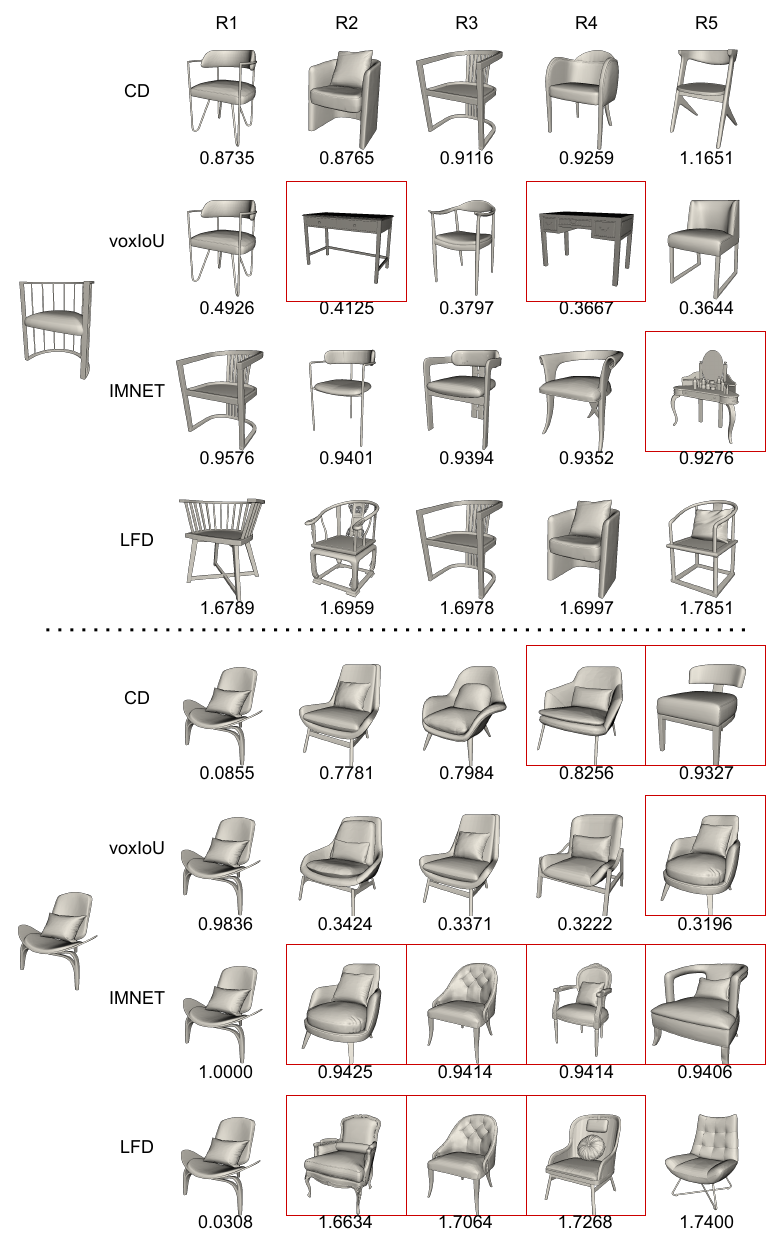}
    \caption{Comparison of shape rankings under different \textbf{shape2shape} metrics. We show top-5 rankings excluding the ground-truth. The red box indicates a shape is noticeably different from the query shape due to distinguishable geometry or even wrong category.  Note that for the distance metrics (CD, LFD), lower scores indicate shapes that are more similar, and that for the similarity based metrics (voxIOU, IMNET), higher scores indicate shapes that are more similar.  We see that both voxIoU and IMNET sometimes retrieve shapes that are of the wrong category.}
    \label{fig:s2s_metric_comparison}
\end{figure}

Besides point clouds, 3D shapes can be represented using other formats such as voxels, multi-view images and neural descriptors. Therefore, we assess the potential of metrics based on such alternative 3D representations.
\textbf{VoxIoU} computes the intersection-over-union between voxelizations of two shapes.
We generate solid voxels at $128^3$ resolution using binvox\footnote{\href{https://www.patrickmin.com/binvox/}{www.patrickmin.com/binvox}}. Similar to point cloud--based metrics, this metric is sensitive to scale and alignment issues.
For neural shape descriptors, we use \textbf{IMNET}~\cite{chen2019learning} as the shape encoder due to its simple but effective architecture.
We train an IMNET model on 3D-FUTURE shapes from the first 1,500 3D-FRONT~\cite{fu20203dfront} scenes: 1083 unique objects including 312 chairs, 177 beds, 351 sofas and 253 tables. We compute cosine similarity between two normalized IMNET embeddings. Lastly, we represent each 3D shape as light-field descriptor, \textbf{LFD}~\cite{chen2003visual}, computed from a set of pre-rendered binary masks. Assuming all 3D shapes are normalized and centered in the same canonical orientation, 200-view renderings are obtained by placing cameras on vertices of 10 randomly rotated dodecahedrons. We use averaged L1 distances over all views to measure shape similarity. In general, these three shape2shape metrics produce shape rankings with comparable quality in a way that they may rank less similar shapes higher (see voxIoU in the 1st example, IMNET and LFD in the 2nd example in \Cref{fig:s2s_metric_comparison}).
As the IMNET model is trained on a limited number of 3D shapes, we believe the neural descriptor-based metric has potential for producing better rankings with more training data and more advanced neural shape descriptor models.
Taking into account the computational simplicity and robustness, we choose LFD over voxIoU and IMNET.

\subsection{View-dependent Metrics}
\label{sec:supp-metrics-view}

\begin{figure}
    \includegraphics[width=\linewidth,trim={0 0.3cm 0 0.3cm},clip]{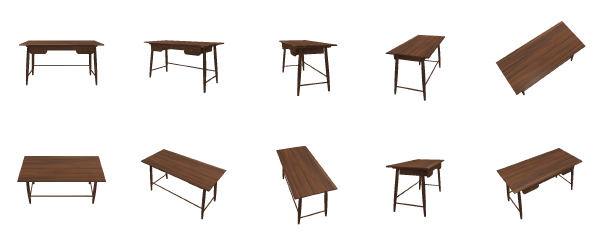}
    \caption{Example of 10 randomly sampled camera viewpoints for one object. 
    We use these viewpoints for analyzing the quality of the different view-dependent metrics.}
    \label{fig:10_random_views}
\end{figure}

\begin{figure}    \includegraphics[width=\linewidth]{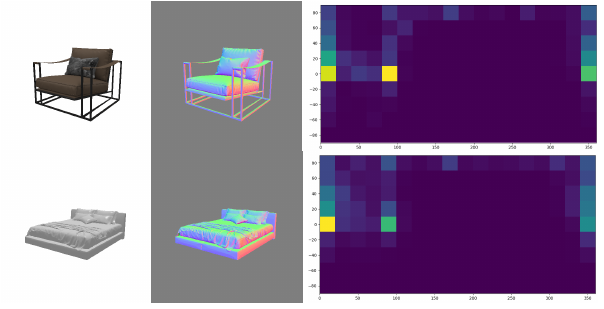}
    \caption{The nIOU metric may score two very dissimilar shapes as highly similar.  Here we show an example where two 2D normal histograms of two different shapes are similar. We set the bin size to 20 for this example.}
    \label{fig:normal_histograms}
\end{figure}

\begin{figure}
    \includegraphics[width=\linewidth,trim={0 0.2cm 0 0.2cm},clip]{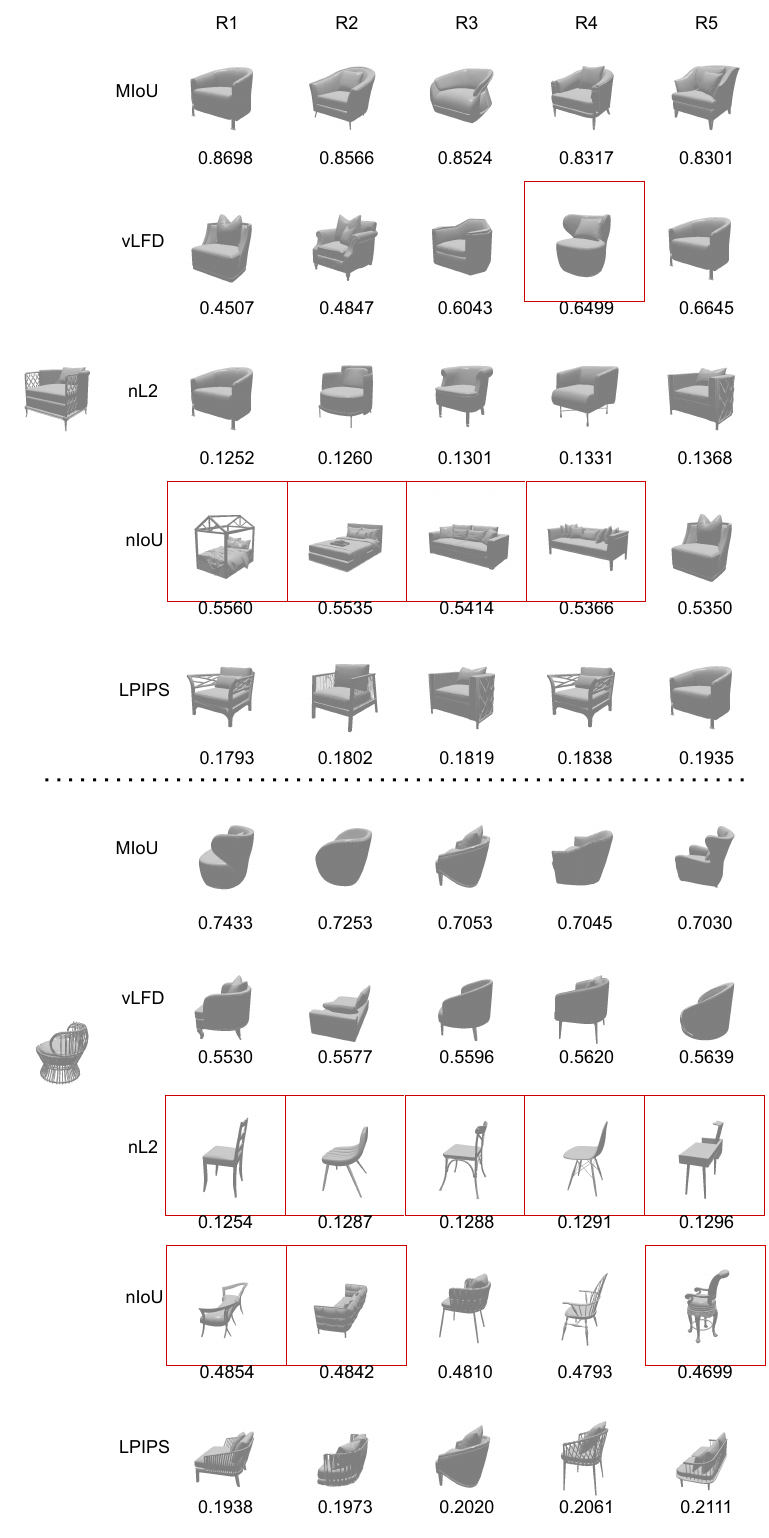}
    \caption{Comparison of shape rankings under different \textbf{view-dependent} metrics. We show top-5 rankings excluding the ground-truth. The red box indicates a shape is noticeably different from the query shape due to distinguishable geometry or even wrong category. 
    Note that for the distance metrics (vLFD, nL2, LPIPS), lower scores indicate shapes that are more similar, and that for the similarity based metrics (nIoU), higher scores indicate shapes that are more similar.  We see that both voxIoU and IMNET sometimes retrieve shapes that are of the wrong category.  We find that both nL2 and nIoU are rather poor at matching ground-truth shape.}    \label{fig:view_metric_comparison}
\end{figure}

We compare the view-dependent metrics we use in the main paper: mask IOU (MIoU), and single-view LFD L1 distance (vLFD), LPIPS~\cite{zhang2018unreasonable} with two view-dependent normal based metrics inspired by~\cite{kuo2021patch2cad}, normal $L2$ distance (nL2) and normal IoU (nIoU). To evaluate the ranking quality of the view-dependent metrics, the inputs are either mask images, neutral renderings or normal maps for each 3D shape rendered under 10 randomly sampled camera viewpoints using PyTorch3D (see \Cref{fig:10_random_views}).
% In Patch2CAD, for a patch of normals, the authors compute a self-similarity histogram over all pairwise angular distances of the normals in the patch and then measure IoUs between different patches to obtain an orientation-independent metric. In our case, for image mask to shape retrieval evaluation, we only need to consider one pose.  Thus, we do not need to compute self-similarity histogram over all pairwise angular distances.
%\qirui{We don't compute the full histogram because (1) shape views for retrieval are all rendered under the same pose, so whether orientation-independent or not doesn't affect the result. (2) it is computation-consuming to obtain pairwise angular distance given relatively high resolution.} 
% Instead, we consider two simplified and computationally efficient metrics that measures L2 difference between normals (nL2) and IoU over histogram of normals (nIoU). 

For \textbf{MIoU} and \textbf{vLFD}, only the binary segmentation mask is used to rank shapes by computing mask IoU and single-view LFD $L1$ distance respectively.
% \textbf{Normal IoU and Distance.} 
Since both MIoU and vLFD mainly measure similarity between object silhouettes without considering geometry variants inside the object, we consider measuring view-dependent normal similarity derived from patch based normal similarity to determine patch-wise matches during training of Patch2CAD~\cite{kuo2021patch2cad}. In Patch2CAD, for a patch of normals, the authors compute a self-similarity histogram over all pairwise angular distances of the normals in the patch and then measure IoUs between different patches to obtain an orientation-independent metric. In our case, for image mask to shape retrieval evaluation, we only need to consider one pose. 
% Thus, we do not need to compute self-similarity histogram over all pairwise angular distances. 
Thus, we consider two simplified and computationally efficient metrics that measures L2 difference between normals (nL2) and IoU over histogram of normals (nIoU). 
% To compute these metrics, we first render normal maps for each 3D shape using the sampled 10 camera viewpoints.
We then define \textbf{nL2} and \textbf{nIoU} where the former is the average $L2$ distance between normals over all pixels on the object, and the latter computes the IoU between 2D normal histograms of two shapes. To build the normal histogram, we represent each normal vector as two angles for azimuth and elevation such that they fall into two bins partitioned on azimuth and elevation separately. The IoU between two 2D normal histograms is obtained by taking the average of bin-wise IoUs over all bins. We set the bin size to 10 for both azimuth and elevation. As normal IoU is invariant to actual positions of normals, two different shapes can share similar 2D normal histogram pattern (see \Cref{fig:normal_histograms}). For perceptual similarity measurement \textbf{LPIPS}, we use neutral renderings (white colored shapes) to reduce the effect of textures and resize images to 100x100 for passing into a pretrained VGG model \footnote{\href{https://github.com/richzhang/PerceptualSimilarity}{github.com/richzhang/PerceptualSimilarity}}.
% Similar to the post-processing on mask renderings, we remove invisible parts of normal maps due to object occlusion. For each pair of GT normal and predicted normal, we compute their distance by averaging all pairwise angular distances of the normals the visible area. 
We show two views of two query shapes in \Cref{fig:view_metric_comparison}.

We observe that MIoU and vLFD measure more coarse-grained and overall geometry consistency with the query shape since they only get access to mask information.
We find that nL2 can produce competitive rankings against MIoU and vLFD when the shape possess relatively simple structure. When the local structure of the target shape becomes complicated, both nL2 and nIoU fail to return meaningful shape rankings.
LPIPS~\cite{zhang2018unreasonable} is more robust to capturing fine-grained structure similarity (see thin slat structure presented in top-5 rankings of LPIPS in \Cref{fig:view_metric_comparison}) given different views of shapes.

\section{Additional Results}
\label{sec:supp-results}
In this section, we present additional results. 
% We show that by training a model with the MOOS data, it is possible to fine-tune a model on Pix3D data with considerably less training data (\Cref{sec:supp-results-fewshot}).  
We conduct an ablation study (\Cref{sec:supp-results-ablation}) that shows that some of the proposed techniques such as color transfer for data augmentation and unique shape mining is not necessary for good image-to-shape retrieval.  Finally, we present more qualitative results on Pix3D~\cite{sun2018pix3d} and ScanNet~\cite{dai2017scannet} frames (\Cref{sec:supp-results-qualitative}), focusing on examples where the top retrieved shape does not match the ground truth model exactly.   

% \subsection{Few-shot Fine-tuning}
% \label{sec:supp-results-fewshot}

% \input{tables/few_shot}
% \begin{table}
% \centering
% \resizebox{\linewidth}{!}
% {
% \begin{tabular}{@{}l cccccc@{}}
% \toprule
% %  & \multicolumn{3}{c}{Retrieval Accuracy} & \multicolumn{4}{c}{Shape-wise Metrics} & \multicolumn{6}{c}{View-dependent Metrics} \\
% % \cmidrule(lr){2-4}\cmidrule(lr){5-8}\cmidrule(lr){9-14}
% CMIC & $Acc_1$$\uparrow$ & CD$\downarrow$ & LFD$\downarrow$ & MIoU$\uparrow$ & vLFD$\downarrow$ & LPIPS$\downarrow$ \\
% \midrule
% Baseline & 48.2 & \textbf{0.6496} & 1.4109 & \textbf{0.7712} & \textbf{0.7644} & \textbf{0.3048}  \\
% Crop object & 47.6 & 0.7614 & \textbf{1.3999} & 0.7642 & 0.7852 & 0.3103  \\
% Color Transfer~\cite{lin2021single} & \textbf{48.3} & 0.7051 & 1.4042 & 0.7707 & 0.7694 & 0.3075  \\
% Unique Shape Mining & 43.9 & 0.8085 & 1.5222 & 0.7467 & 0.8493 & 0.3209  \\
% \midrule
% R18 encoders & 47.4 & 0.7966 & 1.4232 & 0.7638 & 0.7822 & 0.3093  \\
% ViT query encoder & 47.7 & 0.7162 & 1.4211 & 0.7690 & 0.7728 & 0.3074  \\
% Multihead attention & 47.4 & 0.8544 & 1.4220 & 0.7568 & 0.8055 & 0.3123  \\
% Stacked attention & 46.8 & 0.8248 & 1.4156 & 0.7554 & 0.8256 & 0.3130  \\
% % \midrule
% \bottomrule
% \end{tabular}
% }
% \caption{Ablations of CMIC~\cite{lin2021single} on Pix3D comparing training techniques and network submodule design choices.  We find that it is not necessary to use color transfer or unique shape mining as was done in the original paper and implementation.}
% \label{tab:ablation_study}
% \end{table}

\begin{table}
\centering
\resizebox{\linewidth}{!}
{
\begin{tabular}{@{}l cccccc@{}}
\toprule
%  & \multicolumn{3}{c}{Retrieval Accuracy} & \multicolumn{4}{c}{Shape-wise Metrics} & \multicolumn{6}{c}{View-dependent Metrics} \\
% \cmidrule(lr){2-4}\cmidrule(lr){5-8}\cmidrule(lr){9-14}
CMIC & $Acc_1$$\uparrow$ & CD$\downarrow$ & LFD$\downarrow$ & MIoU$\uparrow$ & vLFD$\downarrow$ & LPIPS$\downarrow$ \\
\midrule
Baseline & 48.2 & \textbf{0.650} & 1.411 & \textbf{0.771} & \textbf{0.764} & \textbf{0.305}  \\
Crop object & 47.6 & 0.7614 & \textbf{1.400} & 0.764 & 0.785 & 0.310  \\
Color Transfer~\cite{lin2021single} & \textbf{48.3} & 0.705 & 1.404 & 0.771 & 0.769 & 0.308  \\
Unique Shape Mining & 43.9 & 0.809 & 1.522 & 0.747 & 0.849 & 0.321  \\
\midrule
R18 encoders & 47.4 & 0.797 & 1.423 & 0.764 & 0.782 & 0.309  \\
ViT query encoder & 47.7 & 0.716 & 1.421 & 0.769 & 0.773 & 0.307  \\
Multihead attention & 47.4 & 0.854 & 1.422 & 0.757 & 0.806 & 0.312  \\
Stacked attention & 46.8 & 0.825 & 1.416 & 0.755 & 0.826 & 0.313  \\
% \midrule
\bottomrule
\end{tabular}
}
\caption{Ablations of CMIC~\cite{lin2021single} on Pix3D comparing training techniques and network submodule design choices.  We find that it is not necessary to use color transfer or unique shape mining as was done in the original paper and implementation.}
\label{tab:ablation_study}
\end{table}

\subsection{Ablation Study}
\label{sec:supp-results-ablation}

We perform a set of ablation studies for CMIC on Pix3D in terms of training techniques and model architecture as shown in the two subgroups in \Cref{tab:ablation_study}. The baseline is a CMIC model taking as input the whole RGB image and applying several data augmentations including affine transformation, crop, flip and color jitter.

We find that object cropping degrades the performance on Pix3D where images mainly contain one salient object, but improves retrieval given inputs containing multiple objects like \moos. In the CMIC paper, \citet{lin2021single} proposes color transfer to augment image queries, but our results show that simpler color jitter can achieve nearly the same performance.
The original implementation of CMIC\footnote{\url{https://github.com/IGLICT/IBSR_jittor}} uses unique shape mining when preparing 3D shapes in a batch.
Our experiments show that actually results in worse performance.

\begin{figure}
    \includegraphics[width=\linewidth]{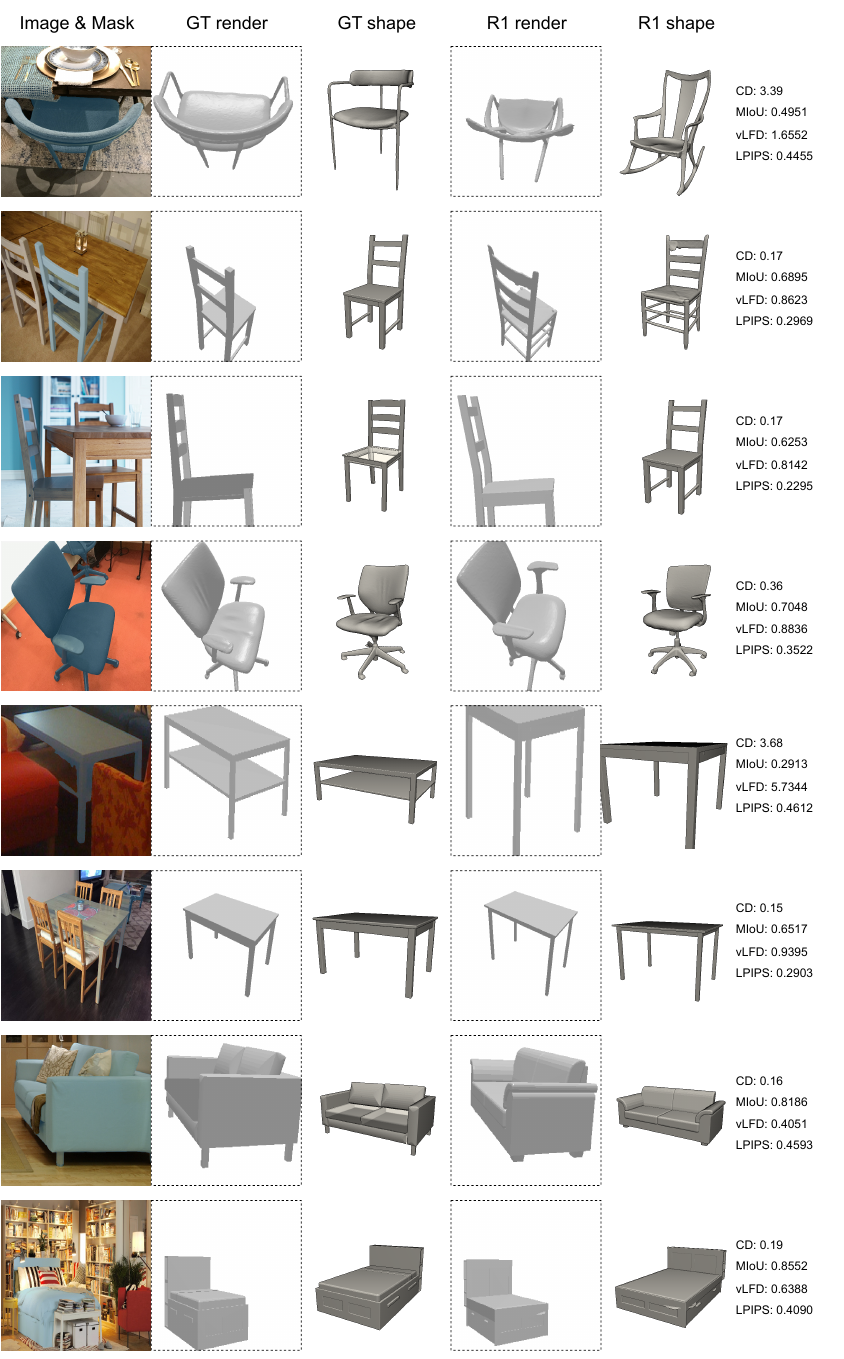}
    \caption{Qualitative results on Pix3D unseen occluded objects with CMIC-MOOS-ft model fine-tuned on Pix3D, including view-dependent and complete renderings of both GT shape and top@1 retrieved shape. For all these examples, the top@1 retrieved shape is not the exact same model as the ground truth model.}
    \label{fig:pix3d_detailed_examples}
\end{figure}

\begin{figure}
    \includegraphics[width=\linewidth]{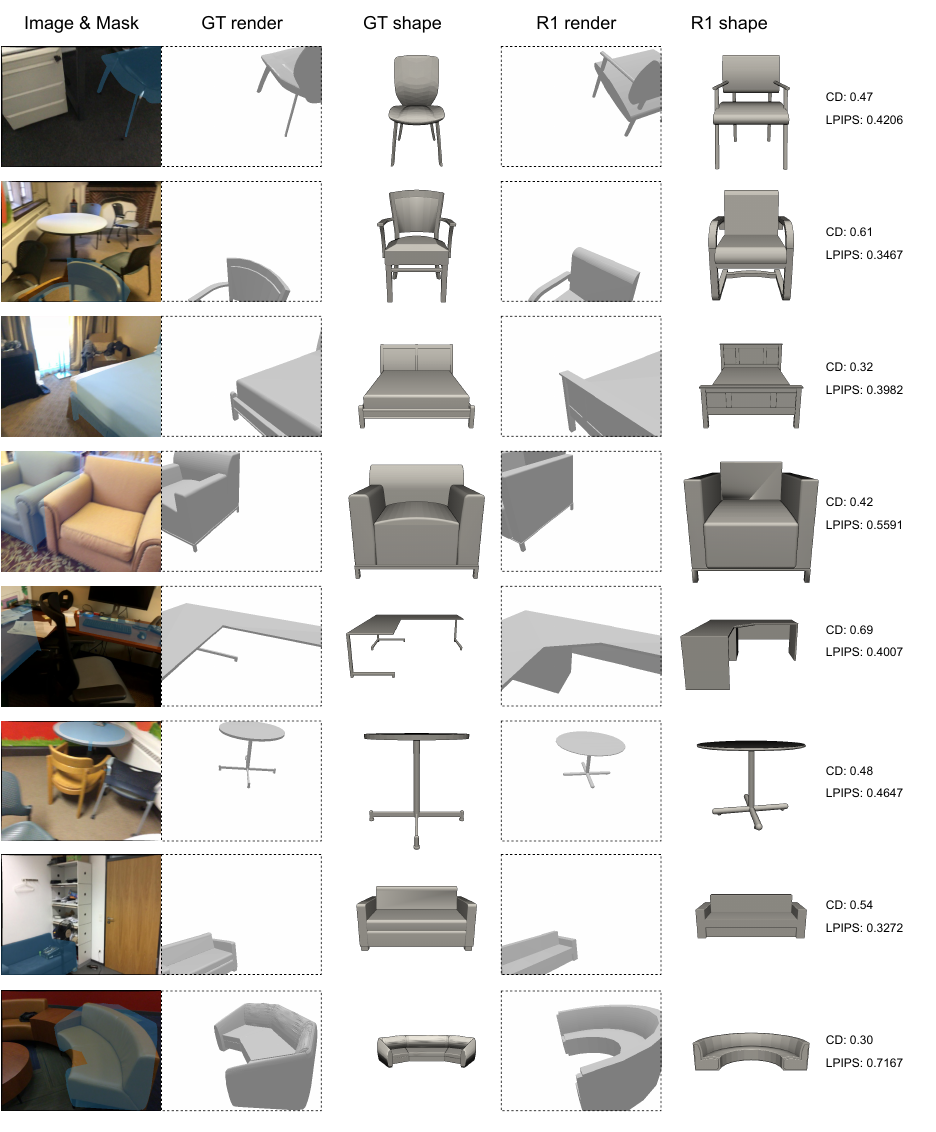}
    \caption{Qualitative results on Scan2CAD unseen objects with CMIC-MOOS-ft model fine-tuned on Scan2CAD, including view-dependent and complete renderings of both GT shape and top@1 retrieved shape. For all these examples, the top@1 retrieved shape is not the exact same model as the ground truth model. However, we see that the retrieved models are mostly plausible matches to the image.}
    \label{fig:scan2cad_detailed_examples}
\end{figure}

We also experiment with various settings for different components of CMIC, including using R18 for both encoders, using ViT~\cite{dosovitskiy2020image} as the query encoder, and replacing the dot production attention with multi-head attention or stacked attention blocks.
However, none of these variants show promising signals of performance improvement. For the ViT encoder, we use a small ViT model pretrained on ImageNet21K~\cite{krizhevsky2012imagenet} with patch size of 32.
Similar to ResNet-based encoders, we add a linear projection layer to process the extra mask channel and an MLP layer to embed feature into the embedding space of dimension 128. For multi-head attention, we use a common multi-head attention block from Transformer~\cite{vaswani2017attention} with 6 heads and feature dimension of 384. For stacked attention, we stack two multi-head attention blocks together.

\subsection{Qualitative Results}
\label{sec:supp-results-qualitative}

We show detailed qualitative results on Pix3D~\cite{sun2018pix3d} and Scan2CAD~\cite{avetisyan2019scan2cad} (see \Cref{fig:pix3d_detailed_examples,fig:scan2cad_detailed_examples}), including input RGB images with masks, view-dependent metrics and complete renderings of ground-truth and top@1 retrieved shapes.  In this qualitative results, we focus on showing examples where the retrieve shape is not the exact same model as the ground-truth.  However, we see that in many cases, the retrieved shape is quite similar to the ground truth.
% \todo{For qualitative results on Pix3D and Scan2CAD, we will provide object masks, complete ground truth and retrieved shapes} 
\fi

\end{document}